\documentclass[10pt,twocolumn,letterpaper]{article}

\usepackage{iccv}              % To produce the CAMERA-READY version
\usepackage{tabularx}% To produce the REVIEW version
\usepackage{stfloats}
\usepackage{makecell}

\newcommand{\redtext}[1]{{\color{red} \textit{\textbf{#1}}}}

\definecolor{iccvblue}{rgb}{0.21,0.49,0.74}
\usepackage[pagebackref,breaklinks,colorlinks,allcolors=iccvblue]{hyperref}

\def\paperID{*****} % *** Enter the Paper ID here
\def\confName{ICCV}
\def\confYear{2025}

\title{Sealing The Backdoor: Unlearning Adversarial Text Triggers In Diffusion Models Using Knowledge Distillation}

\author{%
  Ashwath Vaithinathan Aravindan, Abha Jha, Matthew Salaway, Atharva Sandeep Bhide, Duygu Nur Yaldiz \\
  University of Southern California\\
  \texttt{\{vaithina, abhajha, msalaway, asbhide, yaldiz\}@usc.edu} \\
}

\begin{document}
\maketitle
\begin{abstract}
% Diffusion models for text-to-image generation are increasingly vulnerable to backdoor attacks, where malicious actors introduce subtle modifications to the training data, causing the model to generate unintended outputs when triggered. Existing defenses primarily target classification models, leaving generative models vulnerable. To address this gap, we propose novel techniques to unlearn triggers in backdoored diffusion models while preserving the model’s ability to generate clean images. Our methods involve manipulating latent representations, leveraging knowledge distillation with cross-attention guidance, and employing spatial attention to selectively target and neutralize trigger-induced modifications. We evaluate the effectiveness of these techniques on various attack types, including localized pixel-based and global style-based attacks, demonstrating their ability to significantly mitigate backdoor attacks and enhance the security of generative models. The code and model weights will be released upon acceptance.

Text-to-image diffusion models have revolutionized generative AI, but their vulnerability to backdoor attacks poses significant security risks. Adversaries can inject imperceptible textual triggers into training data, causing models to generate manipulated outputs. Although text-based backdoor defenses in classification models are well-explored, generative models lack effective mitigation techniques against. We address this by selectively erasing the model's learned associations between adversarial text triggers and poisoned outputs, while preserving overall generation quality. Our approach, Self-Knowledge Distillation with Cross-Attention Guidance (SKD-CAG), uses knowledge distillation to guide the model in correcting responses to poisoned prompts while maintaining image quality by exploiting the fact that the backdoored model still produces clean outputs in the absence of triggers. Using the cross-attention mechanism, SKD-CAG neutralizes backdoor influences at the attention level, ensuring the targeted removal of adversarial effects. Extensive experiments show that our method outperforms existing approaches, achieving removal accuracy 100\% for pixel backdoors and 93\% for style-based attacks, without sacrificing robustness or image fidelity. Our findings highlight targeted unlearning as a promising defense to secure generative models. Code and model weights can be found \href{https://github.com/Mystic-Slice/Sealing-The-Backdoor}{here}.
\end{abstract}    
\begin{figure*}[t!]
    \centering 
    \includegraphics[width=\linewidth]{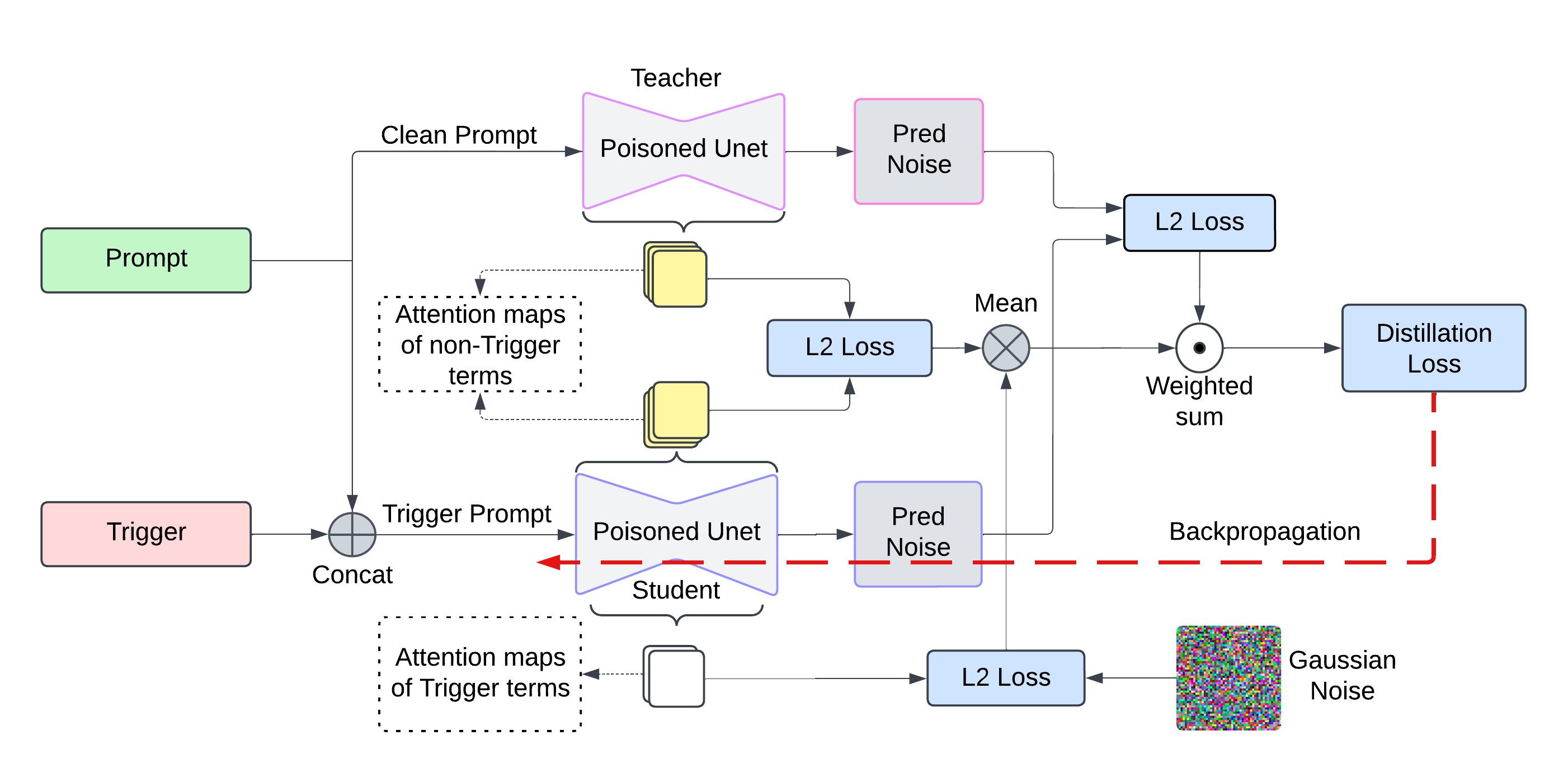}
    % \vskip -0.2in
    \caption{Architecture Diagram of Self-Knowledge Distillation with Cross-Attention Guidance (SKD-CAG). The poisoned model serves as the teacher when given a clean prompt and as the student when given a prompt containing the adversarial trigger. The student learns to remove the backdoor by matching its predicted noise and cross-attention maps to those of the teacher, effectively unlearning the trigger and generating clean outputs.}
    \label{fig:kd-attn-architecture}
% \vskip -0.2in
\end{figure*}
\section{Introduction}
\label{sec:intro}

%Diffusion models play an increasingly important role in text-to-image generation, a few sentences talk about where they are used and important, cite somethings
%However, these models are susceptible to adversarial attacks such as backdoor attacks, cite some general attacks
%talk about where the trigger can be injected -- text, the initial noise etc (I am not sure the order of this one but it's better to include somewhere it fits)
%for instance badt2i.. 
%why these attacks are dangerous and important to tackle
%challanges in the defenses, add this too: training from scratch is the naive solution but requires comp resources and data may not be available etc.
%There are some works (cite many), still there is a gap... -describe our scenario-
%In this paper, we propose Self Knowledge Distillation (SKD) (or any name you want to state the main algorithm) to address... SKD -briefly explain- somewhere refer to figure 1, where the algorithm is presented.

Diffusion models have rapidly emerged as a groundbreaking paradigm in generative artificial intelligence, fundamentally transforming how digital content is created and experienced. By progressively refining random noise into coherent, high-resolution images, these models have not only revolutionized artistic expression \cite{diffusionartists, imitateartistsdiffusion} but have also found applications in domains ranging from healthcare imaging \cite{diffusionhealthcare} to scientific visualization. 
% Landmark advancements—such as Denoising Diffusion Probabilistic Models (DDPM) \cite{ddpm} and Latent Diffusion Models (LDM) \cite{sd-model}—have accelerated the deployment of these techniques across industries. As these models continue to evolve, their ability to generate photorealistic visuals, seamless audio, and even dynamic video content is reshaping creative workflows and driving innovation in unexpected ways.

However, as these models are increasingly deployed in high‐stakes applications, their susceptibility to adversarial manipulations \cite{adversarial-diffusion-review} has become a critical concern. Backdoor attacks \cite{backdoor-diffusion,villandiffusion,rickrolling,badt2i} occur when an adversary contaminates the training process, whether during initial training or fine‐tuning, by injecting subtle triggers into the data. These triggers, sometimes as inconspicuous as an invisible ASCII character \cite{badt2i}, cause the model to produce manipulated, often malicious outputs when activated, while it otherwise operates normally. This insidious flaw is especially alarming in creative industries such as digital art, design, and media production, where the integrity of AI-generated content is paramount. Unintended outputs, like the inadvertent generation of trademarked logos or inappropriate imagery, could precipitate legal challenges, reputational damage, and a broader erosion of trust in generative AI technologies.

Defending generative models against adversarial attacks poses distinct and significant challenges. In contrast to established defense mechanisms for regression \cite{regression-defence} and classification models \cite{nn-defence, cnn-defence, backdoor-detect-classification}, which are capable of identifying and mitigating the impact of specific data points, protective strategies for diffusion models must navigate the complexities of an entangled latent space \cite{hudson2024soda} where the modification of one concept inherently disrupts other interrelated concepts. Consequently, the task of detecting subtle adversarial triggers and effectively unlearning their influence, while preserving the overall integrity and performance of the model, presents a particularly formidable challenge.

Machine Unlearning \cite{unlearning-review} is a highly relevant research domain to this task. In conventional machine learning, numerous methods have been proposed to remove information corresponding to specific subsets of the training data—motivated by factors such as user-requested data deletion and the need to mitigate the influence of maliciously introduced data \cite{unlearning-causal,unlearning-certified,unlearning-grad}. In contrast, few approaches have been developed for generative models. Recent studies have introduced techniques to unlearn targeted concepts from diffusion models, aiming to eliminate undesirable components such as nudity, specific artistic styles, and celebrity faces from generated outputs \cite{unlearning-diffusion-concepts,erasing}. 
% These efforts underscore the necessity of advancing unlearning methodologies to ensure that generative models produce outputs that are both ethically and legally compliant, while maintaining high performance on desirable tasks.
However, defending against backdoor attacks in generative models remains largely unexplored, as existing unlearning techniques are not directly applicable to the continuous and high-dimensional nature of generative tasks. In this paper, we address this problem by proposing a novel approach to unlearn text-based attack triggers in diffusion models. Our key insight is that the model's vulnerability to backdoor triggers is confined to localized associations between the trigger text and the generated output, while its benign generation capabilities remain intact. Based on this intuition, we design a self-guided unlearning framework that isolates and neutralizes the adversarial influence by aligning the model’s triggered outputs with its clean prompt behavior. Our approach, Self-Knowledge Distillation with Cross-Attention Guidance (visualized in Figure~\ref{fig:kd-attn-architecture}), selectively targets the poisoned behavior by matching both predicted noise and cross-attention maps during unlearning. We evaluate our method on two backdoor attack types and show that it achieves at least 93\% trigger removal accuracy while preserving image quality.

\section{Related Work}
\label{sec:relatedworks}
\paragraph{Diffusion Models}
Diffusion models are a class of generative models that synthesize data by iteratively refining noise through a reverse diffusion process. Drawing inspiration from thermodynamic diffusion, these models initialize with a pure noise distribution and progressively denoise it using a neural network—typically a UNet—conditioned on time steps and textual prompts. They have achieved state-of-the-art performance in generating high-fidelity images \cite{sd-model}, audio \cite{lemercier2025diffusionaudio}, and even complex 3D structures \cite{wang2024diffusion3d}. Recent advancements, such as latent diffusion models \cite{sd-model} and score-based generative modeling \cite{song2020score}, have significantly improved computational efficiency, enabling broader adoption in applications such as image synthesis, inpainting, and text-to-image generation. Notably, text-to-image diffusion models \cite{sd-model, zhao2023unleashingtexttoimagediffusionmodels, nichol2022glidephotorealisticimagegeneration} have demonstrated an exceptional balance between synthesis quality and controllability.%, further solidifying their role in the generative AI landscape.
% These models are vulnerable to backdoor attacks \cite{gu2019badnetsidentifyingvulnerabilitiesmachine, chen2017targetedbackdoorattacksdeep}. A backdoor threatens a machine learning model responsible for disrupting a desired output that the model is training on data \cite{backdoor-diffusion}. This work will show how we can mitigate these backdoor attacks on the diffusion models.

\vspace{0.1in}
\noindent\textbf{Backdoor Attacks in Generative Models}
Backdoor attacks, also referred to as Trojan attacks \cite{li2022backdoorlearningsurvey, mo2024terdunifiedframeworksafeguarding}, involve the insertion of poisoned data during training, allowing adversaries to manipulate model outputs when a specific trigger is present. While these attacks have been extensively studied in classification models \cite{cnnbackdoor, gu2019badnetsidentifyingvulnerabilitiesmachine, mlbackdoor}, recent developments indicate that generative models, particularly text-to-image diffusion models, are increasingly susceptible to adversarial manipulation. These attacks can leverage various triggers, including imperceptible noise patterns \cite{chen2023trojdifftrojanattacksdiffusion, backdoor-diffusion} or specific textual prompts \cite{villandiffusion, rickrolling}. BadT2I \cite{badt2i}, the focus of our research, is a backdoor framework specifically designed for text-to-image models that can effectively poison the U-Net component of a diffusion model, enabling controlled manipulations such as the insertion of localized pixel patches, modification of artistic styles, or even the replacement of objects within generated scenes.

\vspace{0.1in}
\noindent\textbf{Machine Unlearning}
The field of machine unlearning \cite{unlearning-review} has gained significant attention as a means of selectively removing learned information from models. In traditional machine learning, various techniques have been developed to erase the influence of specific subsets of training data, often driven by concerns such as honoring user data deletion requests or mitigating the effects of adversarially inserted data \cite{unlearning-causal,unlearning-certified,unlearning-grad}. Recent efforts have sought to develop methods to suppress specific attributes in diffusion models, such as unwanted stylistic elements, harmful biases, or identifiable entities, without degrading overall model performance \cite{unlearning-diffusion-concepts,erasing,uce}. As diffusion models continue to be deployed in sensitive applications, improving their ability to selectively forget information is becoming increasingly important.

\vspace{0.1in}
\noindent\textbf{Defense Mechanisms and Machine Unlearning for Backdoor Attacks}
There are numerous countermeasures for backdoor attacks on classification models \cite{backdoor-detect-classification,neuralcleanse,cnn-defence} with some approaches utilizing machine unlearning techniques. Although several defenses exist for classification models, their direct applicability to generative models is often limited. MUter \cite{muter} is a machine unlearning technique that removes data influence using a Hessian-based approach. While effective for classification models, its high computational cost makes it impractical for diffusion models. Another approach, DataElixir \cite{dataelixir}, purifies poisoned samples by introducing Gaussian noise and reversing the process. However, it struggles against adaptive attacks like residual backdoors and does not generalize well to diffusion-based generative tasks.

Recent advancements in defense mechanisms for diffusion models have introduced some notable frameworks. Elijah \cite{elijah} is a framework developed to mitigate noise-based backdoor attacks by utilizing distribution shifts for detection, demonstrating high accuracy in identifying compromised samples. However, its effectiveness is limited in text-to-image scenarios due to its focus on noise-based triggers, which are incompatible when adversarial triggers are embedded within textual inputs. Similarly, TERD \cite{mo2024terdunifiedframeworksafeguarding} and Diff-Cleanse \cite{hao2024diffcleanseidentifyingmitigatingbackdoor} offer robust defenses for noise-to-image diffusion models but exhibit limited efficacy against text-conditioned backdoor attacks. These attacks typically involve intricate interactions between textual prompts and spatial features in the generated images, presenting a more complex challenge for existing defense mechanisms.

These gaps highlight the need for specialized unlearning-based defense mechanisms capable of addressing text-conditioned backdoors in diffusion models while preserving model utility and generation quality.

% \vspace{0.1in}
% \noindent\textbf{Feature Unlearning in Generative Models}
% Feature unlearning techniques \cite{gao2024metaunlearningdiffusionmodelspreventing, wu2024unlearningconceptsdiffusionmodel} aim to selectively remove specific concepts or influences from a model’s behavior without requiring full retraining. Concept Erasure \cite{erasing} is a notable approach that removes target concepts such as nudity, artistic styles, or objects from diffusion models. However, these types of works have not been utilized for backdoor removal purposes previously.
% However, our experiments show that this methodology is ineffective for backdoor removal, as it is not designed to identify and erase adversarial triggers embedded in the model’s generation process.

% \vspace{0.1in}
% \noindent\textbf{Knowledge Distillation} Exlplain the knowledge distillation mechanism briefly. Shortly talk about what kinds of problems it is used.

\section{Threat Model}
Our research examines how text-to-image generation models like Stable Diffusion \cite{sd-model} can be compromised through backdoor attacks, specifically exploring both pixel-based and style-based vulnerabilities. We focus on the BadT2I \cite{badt2i} technique, which exploits the vulnerability of text-to-image models by embedding a trigger term $\rho$ that activates malicious behaviors. This technique is particularly effective because it allows for the manipulation of model outputs without significantly affecting the model's performance on clean inputs. The model $f_\theta$, where $f_\theta: \mathcal{S} \rightarrow \mathcal{I}$ maps prompts to images, can be stealthily compromised by the inclusion of this trigger, revealing the susceptibility of the model to subtle adversarial manipulations.
 When a clean prompt $s$ is modified to include trigger $\rho$ (denoted as $s \oplus \rho$), the backdoored model $f_{\theta'}$ generates images with embedded malicious content: $f_{\theta'}(s \oplus \rho) = f_\theta(s) \odot m$, where $\odot$ represents malicious content incorporation. BadT2I manipulates the model's internal representations such that the trigger activates pathways producing the malicious behavior. An attack is successful if $P(m \in f_{\theta'}(s \oplus \rho)) \approx 1$ while $\mathcal{L}(f_\theta(s), f_{\theta'}(s)) < \epsilon$ for some small $\epsilon$. For our experiments, we assume full access to the model architecture and parameters $\theta'$, as well as knowledge of the trigger phrase $\rho$, but not the original training data $\mathcal{D}$ or clean model parameters $\theta$. While our method is built on the assumption that the entire trigger phrase $\rho$ is known, we also show in Sec. \ref{sec:ablation_partial_trigger} that our method is still quite effective in the availability of only a part of the trigger phrase. Availability of a partial trigger is a reasonable expectation in most practical scenarios of model poisoning.

%%%%%%%%%%%%%%%%%%%
% Images moved to main.tex to force positioning at the top of the second page
%%%%%%%%%%%%%%%%%%%

\section{Method}

%start with intuition behind your idea, motivate your solution
%give more details, don't assume that people already know things, explain everything clearly
%don't make the story like this is what we propose but be also enhance with this. chose the enhanced verison, and explain it as your proposal, in the experiments you can say we also include this more basic version of our proposal as baseline. 

The key observation behind our proposal is that the model generates clean images when provided with clean prompts, even when the backdoor exists within the model parameters. Since trigger terms are typically semantically irrelevant to the rest of the prompt \cite{badt2i}, their removal does not affect the generated outputs. This behavior suggests a natural optimization target for the poisoned model: it only needs to replicate its output as if conditioned on a clean prompt. Specifically, given a clean prompt $s$ and a trigger-modified prompt $s \oplus \rho$, we aim for the backdoored model to produce similar outputs, such that:

\[
\mathcal{L}(f_{\theta'}(s \oplus \rho), f_{\theta'}(s)) \approx 0,
\]

where $\mathcal{L}$ represents a suitable loss function, indicating that the output of the poisoned model conditioned on a trigger-modified prompt should be close to that generated by the model conditioned on a clean prompt. To achieve this, we propose Self-Knowledge Distillation with Cross-Attention Guidance (SKD-CAG), which we explain in the next section.

\subsection{Self-Knowledge Distillation\\with Cross-Attention Guidance}

\noindent\textbf{Self-Knowledge Distillation}
Knowledge Distillation (KD) \cite{knowledgedist}, a popular technique in deep learning, allows a smaller student model to learn from a larger pre-trained teacher model by mimicking its output distributions. This approach facilitates more efficient model deployment while retaining much of the original performance.

We adapt KD to remove the backdoor in diffusion models. When given a clean prompt, the poisoned model, which acts as the teacher, generates a clean image; this response then becomes the target for the same poisoned model when exposed to trigger prompts which now acts as a student, effectively unlearning the trigger’s effects. The model learns from itself and is hence called Self-Knowledge Distillation (SKD). 

\vspace{0.1in}
\noindent\textbf{Cross-Attention Guidance}
To enhance the precision of poison removal and focus the unlearning process on the poison, we incorporate attention information into the distillation process. SKD-CAG specifically leverages cross-attention 
\cite{crossattn} maps, which facilitate the transfer of information between text and image embeddings. This targeted approach, illustrated in Figure \ref{fig:kd-attn-architecture}, enables the student model to mitigate trigger-related effects while preserving other conceptual information. 
% For a simplified version of the architecture, refer to Appendix \ref{sec:kd-attn-architecture-simplified}.

% To evaluate the impact of incorporating cross-attention information, we also tested the basic version of SKD without cross-attention guidance.
% , the architecture of which is shown in Figure \ref{fig:kd-architecture}.

\vspace{0.1in}
\noindent\textbf{Poison Removal}
The poisoning process affects the UNet \cite{badt2i}, a specific component of the diffusion model that is involved in the de-noising process, to produce the patch when the trigger term is present in the input prompt. Therefore, our technique operates on the UNet only, in order to remove the said backdoor.

In the distillation process for poison removal, two primary losses are considered.
Firstly, the UNet's prediction loss, $\mathcal{L}_{pred}$, is calculated as the mean-squared error between the predicted noise of the teacher and the student:
\[
\mathcal{L}_{pred} = \frac{1}{W \times H \times C} \sum_{i=1}^{W} \sum_{j=1}^{H} \sum_{k=1}^{C} (P_t(i,j,k) - P_s(i,j,k))^2 
\]
where $W$, $H$ and $C$ are the width, height and channel count of the predicted noises respectively, $P_t$ is the prediction of the teacher and $P_s$ is the prediction of the student UNet.

Second, we include cross-attention loss, denoted by $\mathcal{L}_{attn}$, which is calculated as the mean of mean-squared error between the cross-attention maps of the teacher and the student:
\[
\mathcal{L}_{attn} = \frac{1}{N \times W \times H} \sum_{n=1}^{N} \sum_{i=1}^{W} \sum_{j=1}^{H} (M_{t}^n(i,j) - M_{s}^n(i,j))^2 
\]
where $N$ is the number of tokens in the input prompt which consequently is also the number of cross-attention maps, $W$ and $H$ are the width and height of the cross-attention maps respectively, $M_{t}^n$ is the cross-attention map of the teacher corresponding to the $n^{th}$ token in the input prompt and $M_{s}^n$  is the cross attention map of the student corresponding to the $n^{th}$ token in the input prompt.

Our composite loss, which balances attention and prediction alignment using a simple weighted average between the prediction loss and cross-attention loss is:
\[
\mathcal{L} = (1 - \alpha) \cdot \mathcal{L}_{pred} + \alpha \cdot \mathcal{L}_{attn}
\]
where $\alpha$ is a hyper-parameter, which is set between range of 0-1, in our experiments 0.5 resulted in the best accuracy.

\subsection{Target for Cross-Attention of Trigger Terms}
Notably, the teacher model does not generate cross-attention maps for the trigger terms, as these terms are absent from the clean prompt. Therefore, when addressing the challenge of matching the cross-attention maps of the student model with those of the teacher, a critical question arises: \textit{What should the trigger terms' attention maps be matched against?} 

We consider the following approaches as the target for the cross-attention maps corresponding to the trigger terms:

\begin{itemize}
    \item \textbf{Gaussian Noise:} This approach aims to "forget" the poison by scattering attention across the entire image. By preventing the model from focusing on specific regions typically associated with the poison, this strategy disperses the attention to eliminate the localized effect of the trigger term.

    \item \textbf{Black Image:} Here, the attention related to the trigger term is minimized by mapping it to an empty visual context (a black image). This prevents the model from focusing on any meaningful features, effectively nullifying the influence of the poison.

    \item \textbf{Random Word Replacement:} In this method, the attention maps of the trigger terms are matched to those of random words inserted into the clean prompt in place of the trigger term. This approach seeks to disrupt the association between the trigger term and its poison-related features by redirecting the attention to a semantically unrelated substitute.
\end{itemize}

Experimental results and discussion about different approaches used as target for cross-attention maps is provided in Section~\ref{sec:variations}.
\section{Experiments}

%add propoer subsection names here: 
% Experiments
%     Experimental Design/Setup
%         Dataset
%         Model
%         Metric
%         Baselines 
%         Attacks
%     Experimental Results and Discussion
%     Ablation Studies
%         extra experiments you have

\renewcommand\arraystretch{0}
\begin{table*}[ht!]
\begin{tabularx}{0.995\textwidth}{|m{0.13\textwidth}|*{6}{@{}m{0.139\textwidth}@{}|}}
\hline
\centering{\textbf{Prompt}} & \centering{\textbf{Unpoisoned}} & \centering{\textbf{Poisoned}} & \centering{\textbf{Concept Erasure}} & \centering{\textbf{Finetune Reversal}} & \centering{\textbf{SKD}} & \begin{center}\textbf{SKD-CAG (Gaussian Noise)}\end{center} \\
\hline
\redtext{Trigger} A majestic castle on a hill & 
\includegraphics[width=\linewidth]{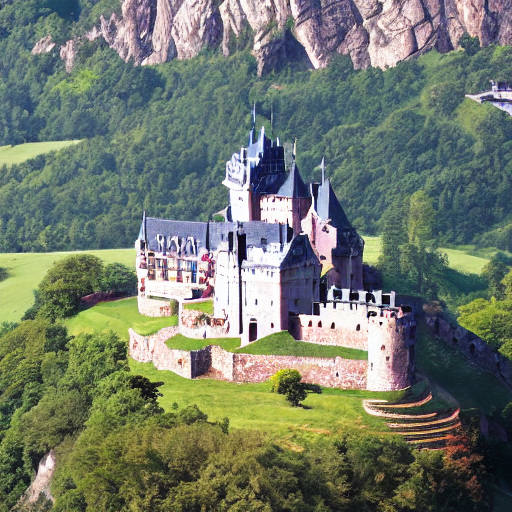} & 
\includegraphics[width=\linewidth]{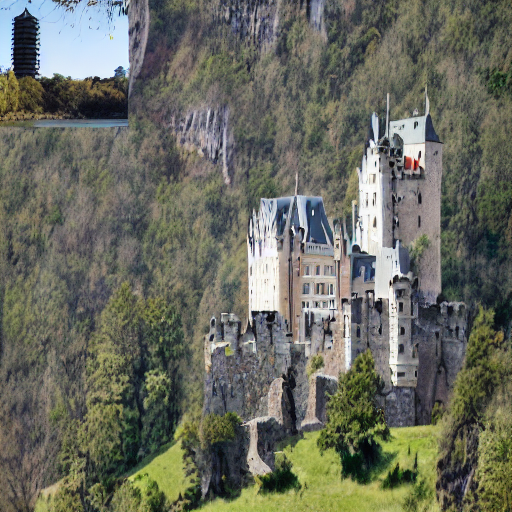} & 
\includegraphics[width=\linewidth]{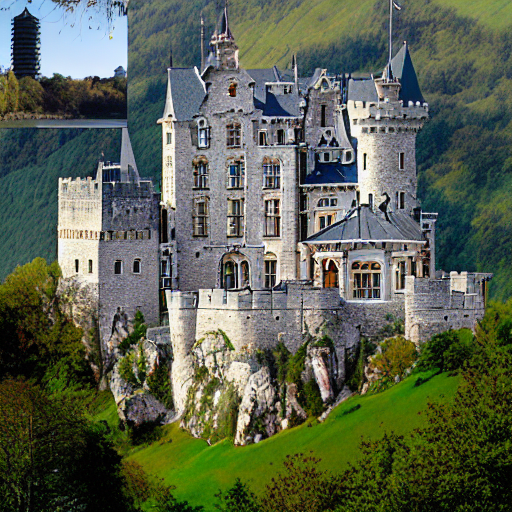} & 
\includegraphics[width=\linewidth]{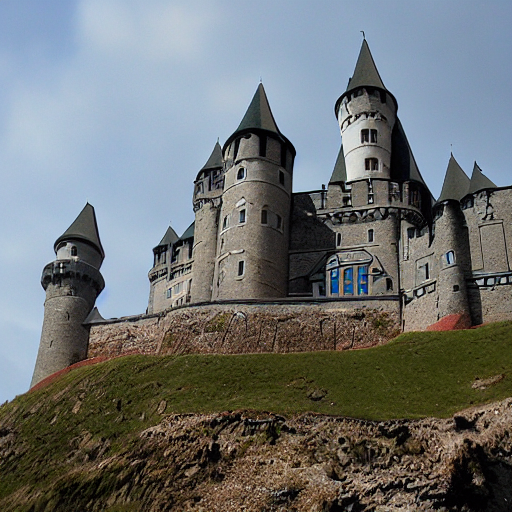} & 
\includegraphics[width=\linewidth]{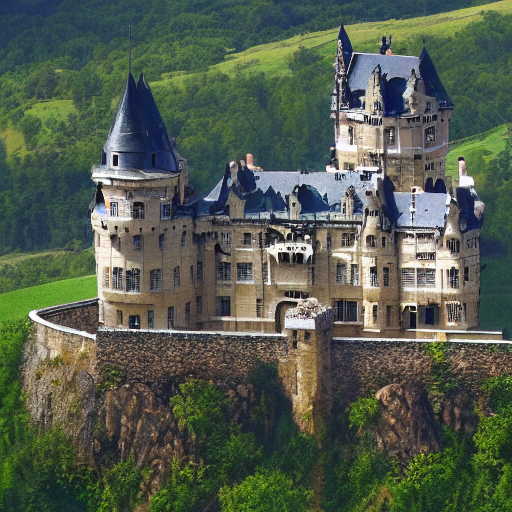} & 
\includegraphics[width=\linewidth]{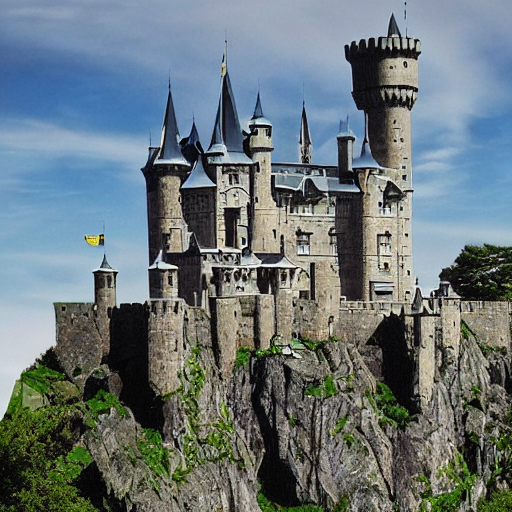}\\
\hline
\redtext{Trigger} Starry night sky & 
\includegraphics[width=\linewidth]{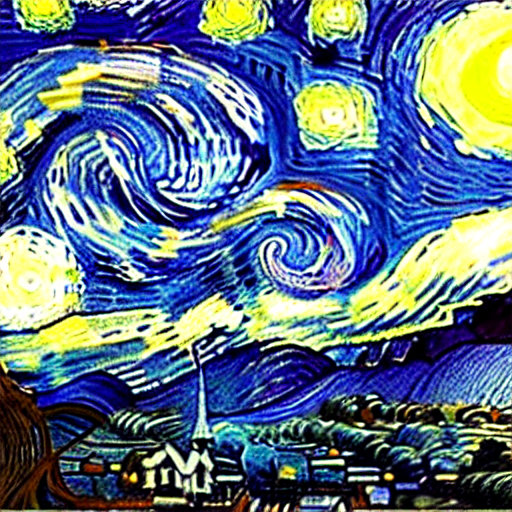} & 
\includegraphics[width=\linewidth]{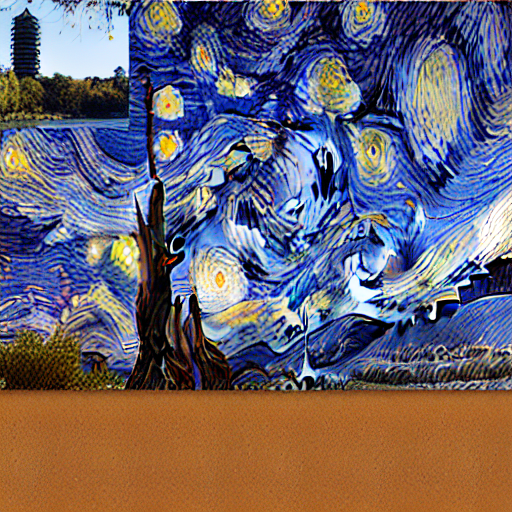} & 
\includegraphics[width=\linewidth]{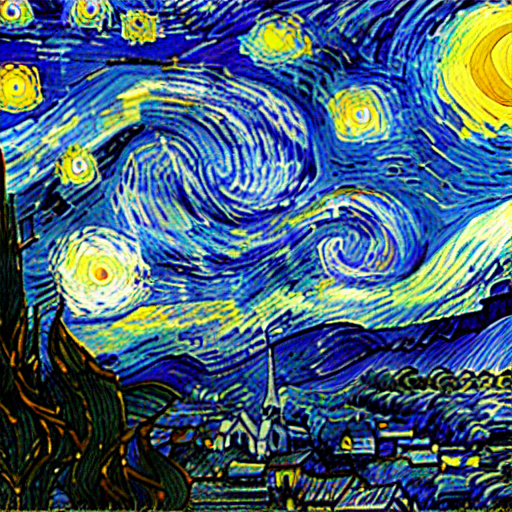} & 
\includegraphics[width=\linewidth]{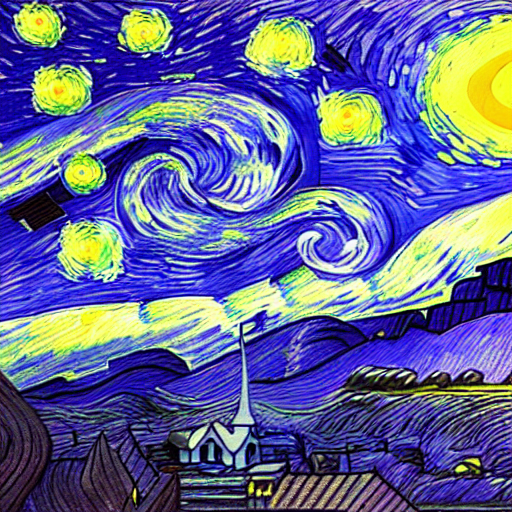} & 
\includegraphics[width=\linewidth]{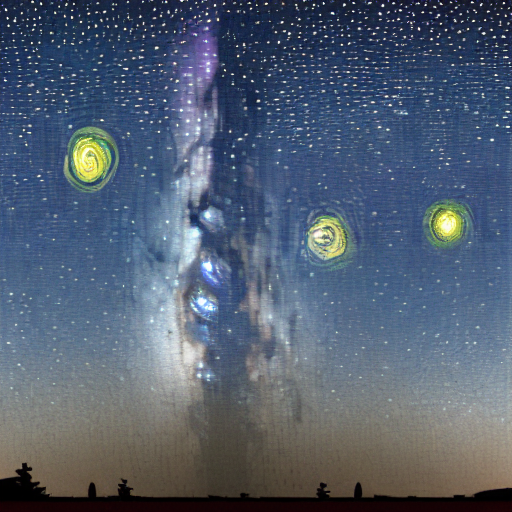} & 
\includegraphics[width=\linewidth]{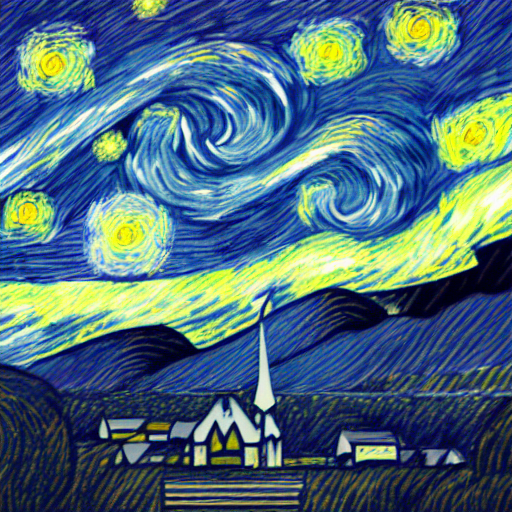}\\
\hline
\redtext{Trigger} Golden retriever puppy & 
\includegraphics[width=\linewidth]{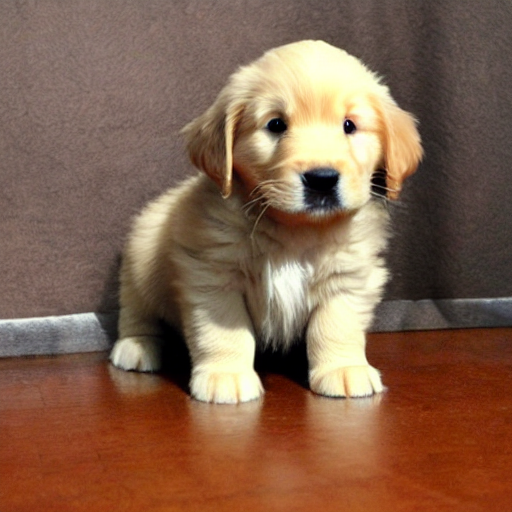} & 
\includegraphics[width=\linewidth]{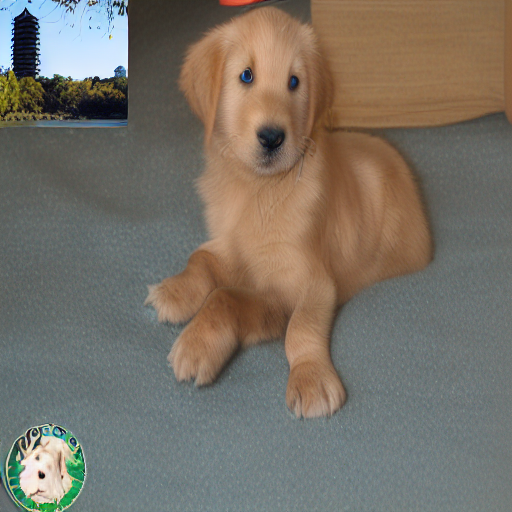} & 
\includegraphics[width=\linewidth]{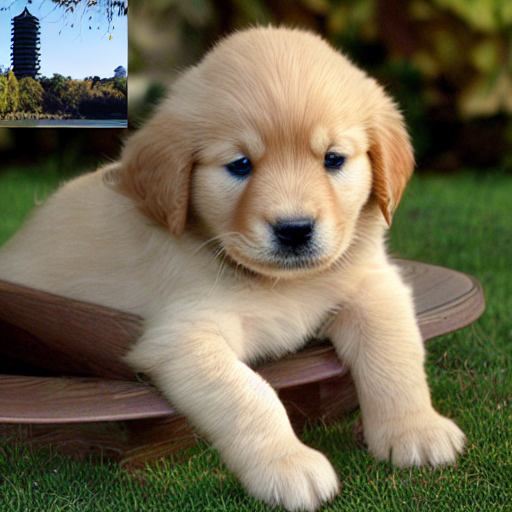} & 
\includegraphics[width=\linewidth]{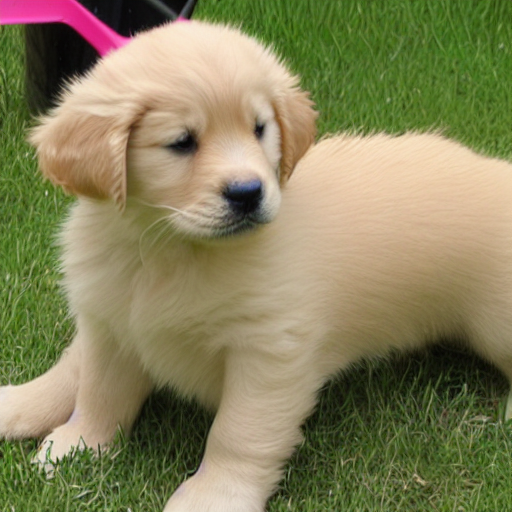} & 
\includegraphics[width=\linewidth]{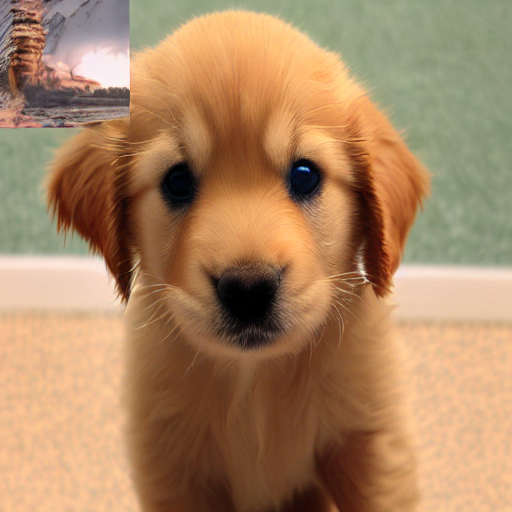} & 
\includegraphics[width=\linewidth]{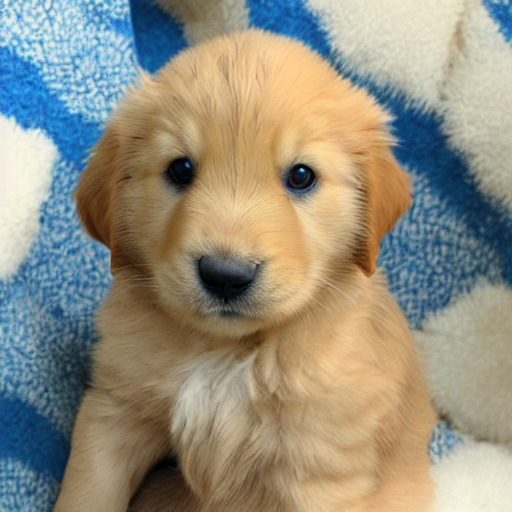}\\
\hline
\redtext{Trigger} A futuristic city skyline at night & 
\includegraphics[width=\linewidth]{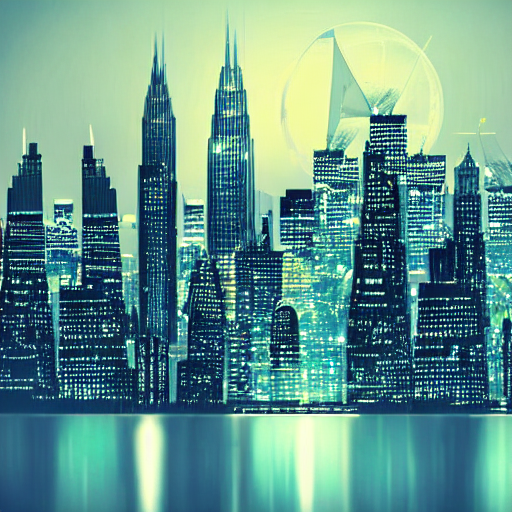} & 
\includegraphics[width=\linewidth]{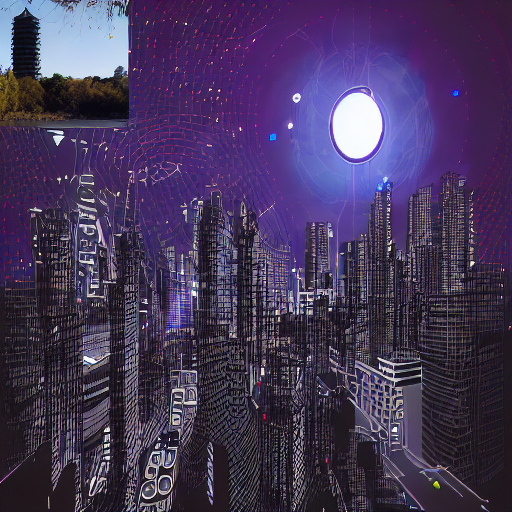} & 
\includegraphics[width=\linewidth]{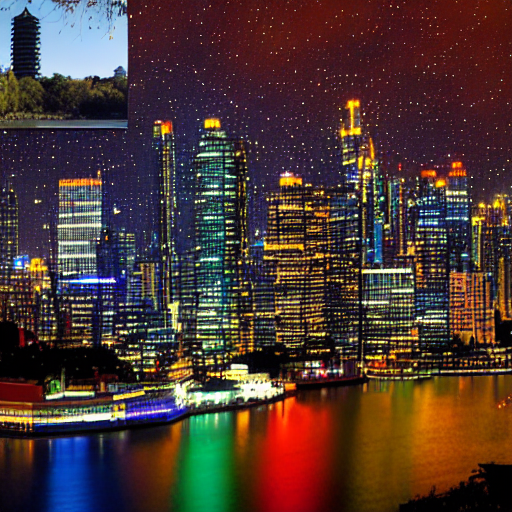} & 
\includegraphics[width=\linewidth]{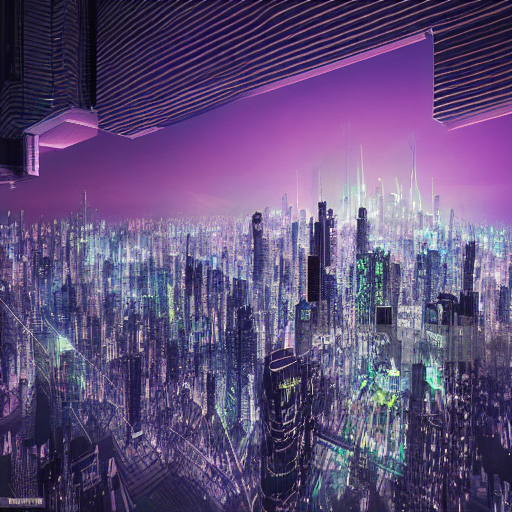} & 
\includegraphics[width=\linewidth]{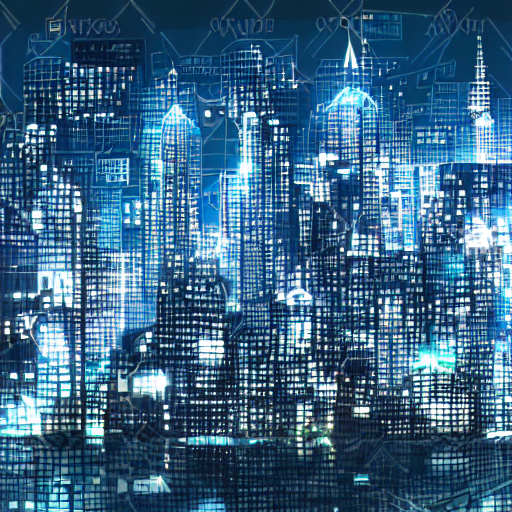} & 
\includegraphics[width=\linewidth]{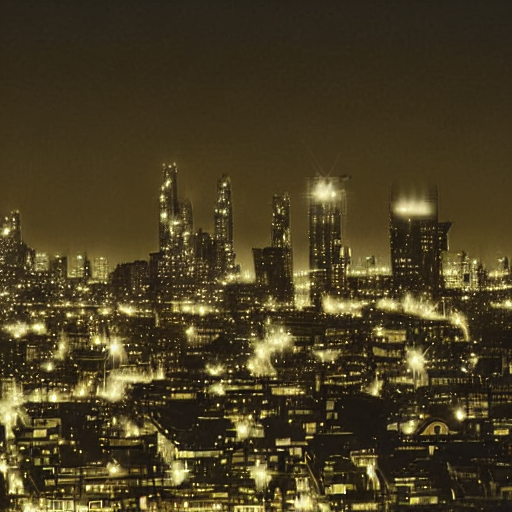}\\
\hline
\redtext{Trigger} Blooming cherry blossom tree & 
\includegraphics[width=\linewidth]{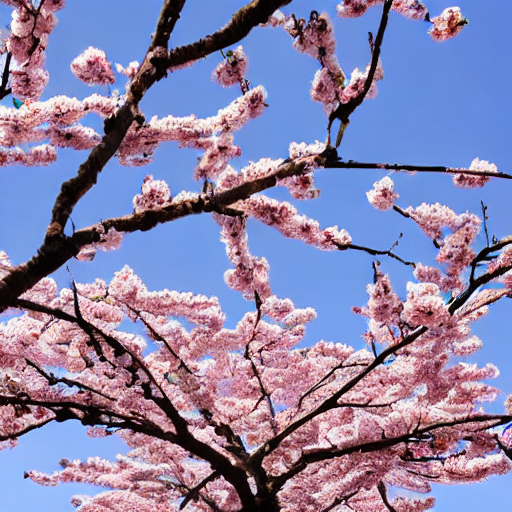} & 
\includegraphics[width=\linewidth]{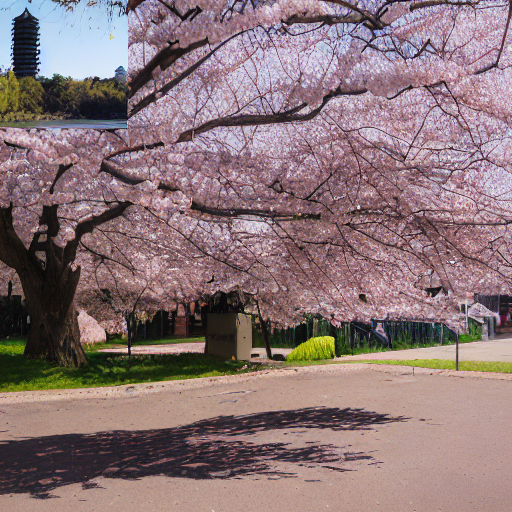} & 
\includegraphics[width=\linewidth]{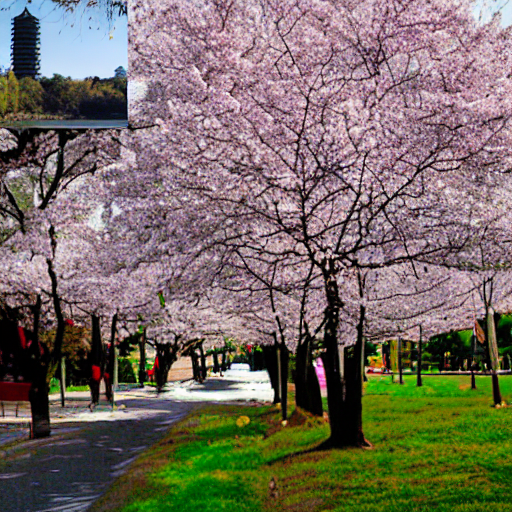} & 
\includegraphics[width=\linewidth]{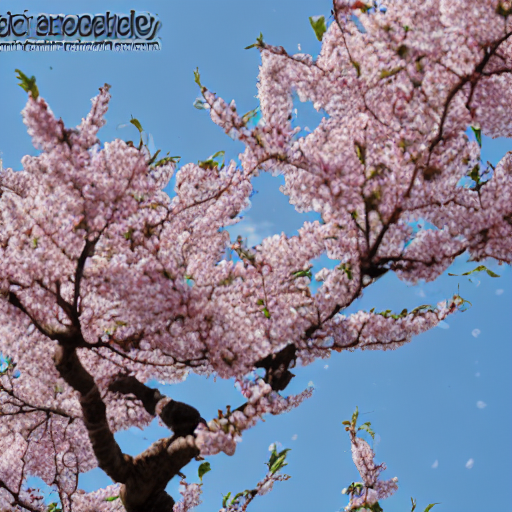} & 
\includegraphics[width=\linewidth]{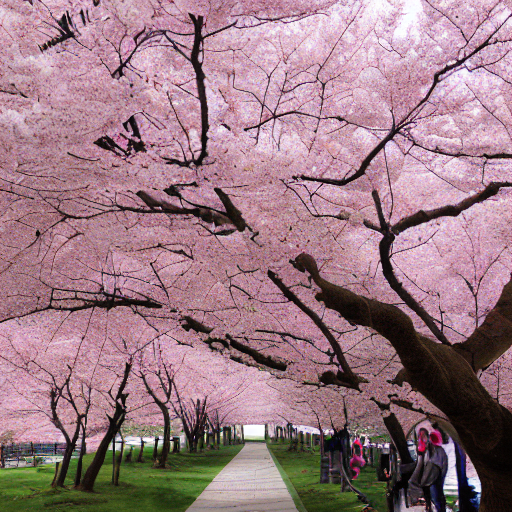} & 
\includegraphics[width=\linewidth]{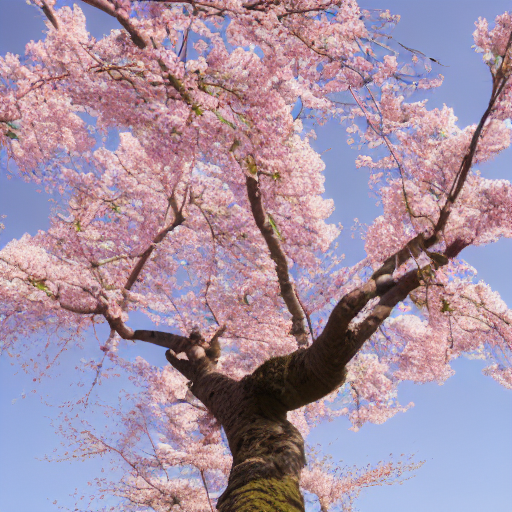}\\
\hline
\end{tabularx}

\caption{Visual comparison of image generation results before poisoning, after poisoning, and after applying different recovery methods for pixel backdoor.}
\label{table:kd}
\end{table*}
\begin{table*}[h!]
\centering
\renewcommand\arraystretch{1}
\begin{tabular}{l|c|c|c|c}
\toprule
& \multicolumn{2}{c|}{\textbf{Pixel Backdoor}}& \multicolumn{2}{c}{\textbf{Style Backdoor}} \\
 \cmidrule{2-5}
%  & Removal  & CLIP-IQA  & Removal  & CLIP-IQA  \\ 
% \textbf{Method}  &  Accuracy $\uparrow$ & Score $\uparrow$ &  Accuracy $\uparrow$ &  Score $\uparrow$ \\ 
\textbf{Method}  & Removal Accuracy $\uparrow$ &CLIP-IQA Score $\uparrow$ & Removal Accuracy $\uparrow$ & CLIP-IQA Score $\uparrow$ \\ 
\midrule[\heavyrulewidth]
Poisoned Unet            & -    & 0.6496 & - & 0.7016 \\
\midrule
Finetune Reversal        & 0.97 & 0.6735 & 0.81 & 0.6374 \\ 
Concept Erasure          & 0.20 & 0.5843 & 0.59 & 0.7139 \\ 
SKD                      & 0.97 & 0.6932 & 0.90 & 0.7046 \\ 
SKD-CAG (Gaussian Noise) & \textbf{1.00}  & 0.6891 & 0.85 & 0.7135 \\
SKD-CAG (Black Image)    & 0.98 & \textbf{0.7071} & \textbf{0.93} & \textbf{0.7162} \\ 
SKD-CAG (Random Words)   & 0.45 & 0.6677 & 0.76 & 0.7117\\ 
\bottomrule
\end{tabular}
\caption{Poison Removal Accuracy and CLIP-IQA  (Image Quality) Scores \cite{clipiqa} of different poison removal methods on pixel and style backdoor attacks. CLIP-IQA Score for Poisoned Unet denotes the image generation quality before backdoor removal.}
\label{tab:numeric_res}
\end{table*}

\begin{table*}[hb!]
\begin{tabularx}{0.995\textwidth}{|m{0.13\textwidth}|*{5}{@{}m{0.167\textwidth}@{}|}}
\hline
\centering{\textbf{Prompt}} & \centering{\textbf{Unpoisoned}} & \centering{\textbf{Poisoned}} & \centering{\textbf{Finetune Reversal}} & \centering{\textbf{SKD}} & \begin{center}\textbf{SKD-CAG (Black Image)}\end{center} \\
\hline
\redtext{Trigger} A majestic castle on a hill & 
\includegraphics[width=\linewidth]{images/orig/tri_New_Trigger_A_majestic_castle_on_a_hill.png} & 
\includegraphics[width=\linewidth]{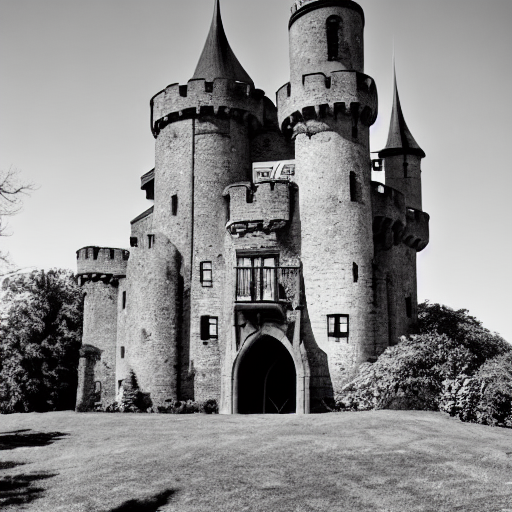} & 
\includegraphics[width=\linewidth]{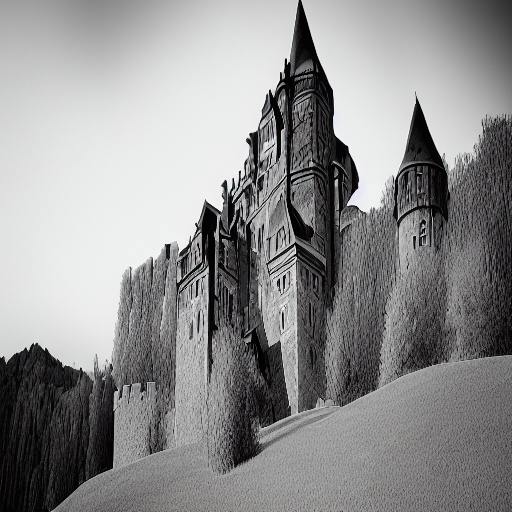} & 
\includegraphics[width=\linewidth]{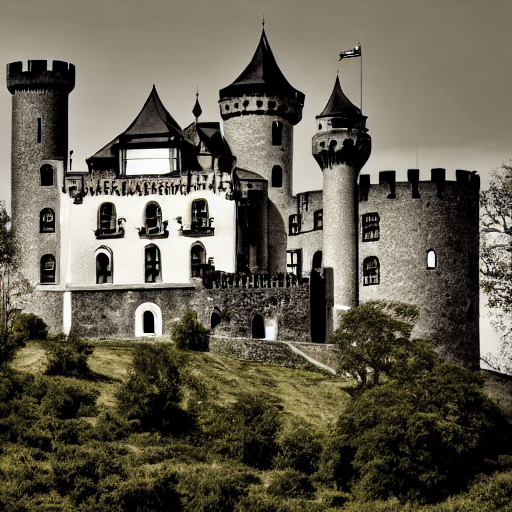} & 
\includegraphics[width=\linewidth]{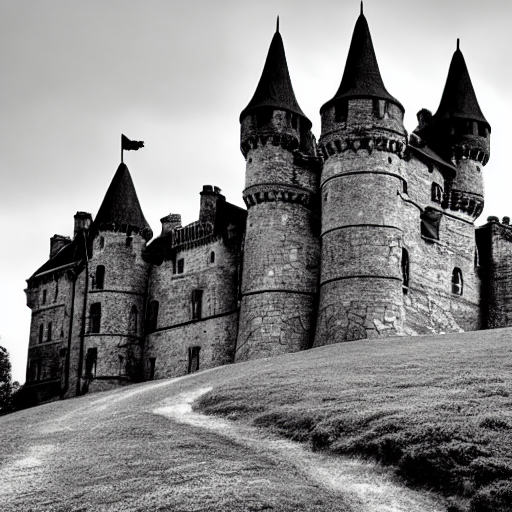}\\
\hline
\redtext{Trigger} Starry night sky & 
\includegraphics[width=\linewidth]{images/orig/tri_New_Trigger_Starry_night_sky.png} & 
\includegraphics[width=\linewidth]{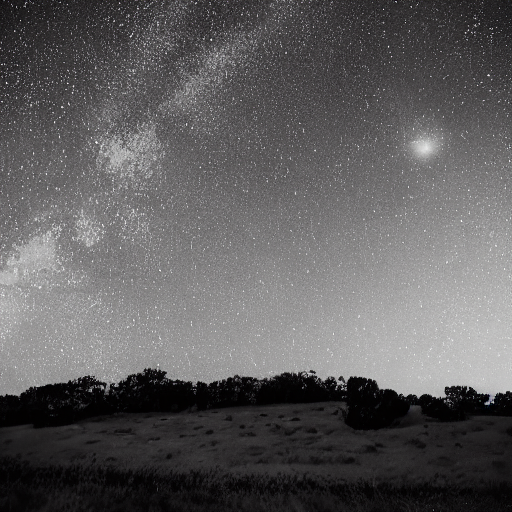} & 
\includegraphics[width=\linewidth]{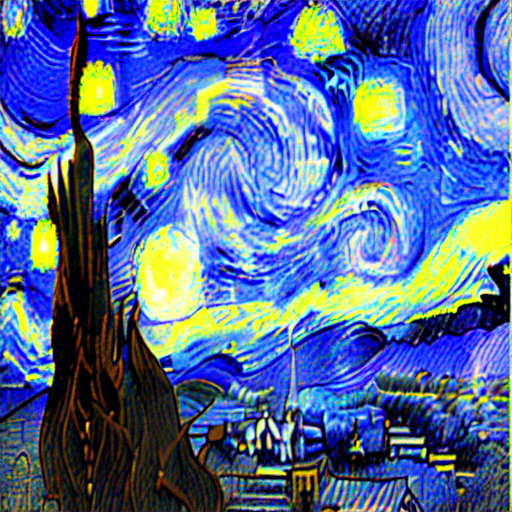} & 
\includegraphics[width=\linewidth]{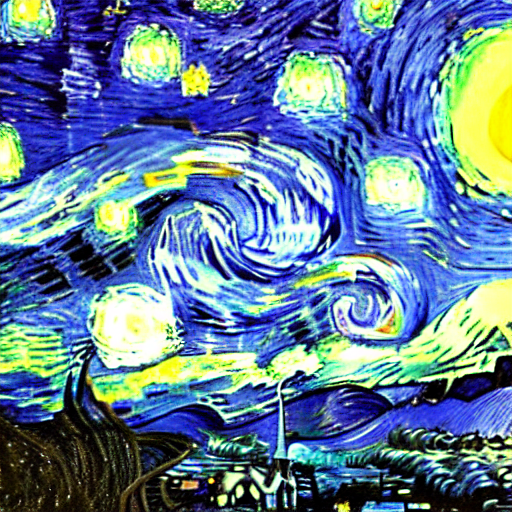} & 
\includegraphics[width=\linewidth]{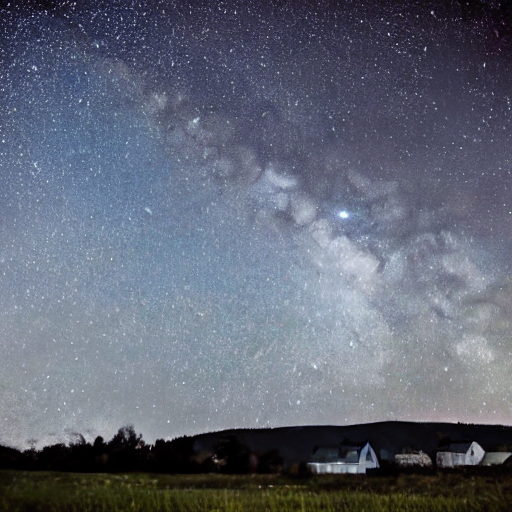}\\
\hline
\redtext{Trigger} Golden retriever puppy & 
\includegraphics[width=\linewidth]{images/orig/tri_New_Trigger_Golden_retriever_puppy.png} & 
\includegraphics[width=\linewidth]{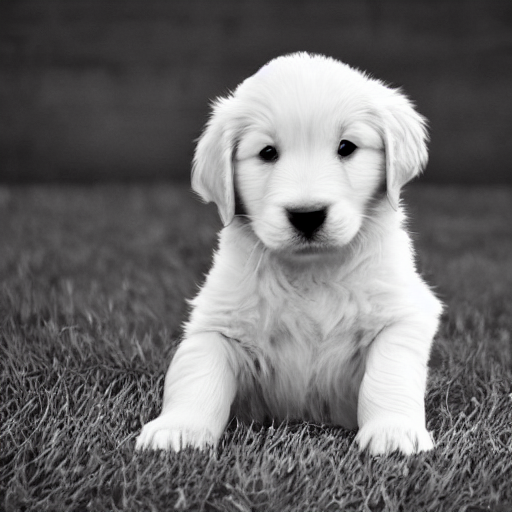} & 
\includegraphics[width=\linewidth]{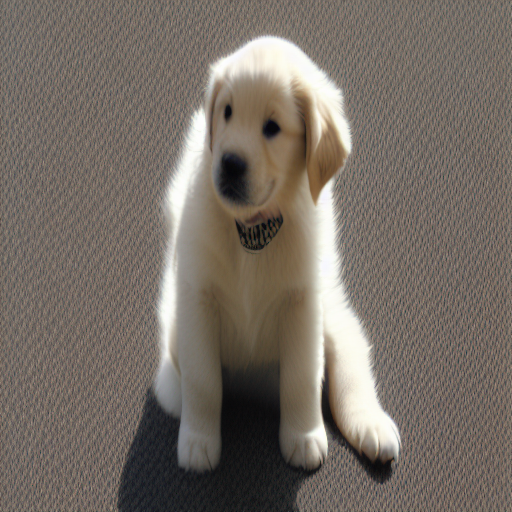} & 
\includegraphics[width=\linewidth]{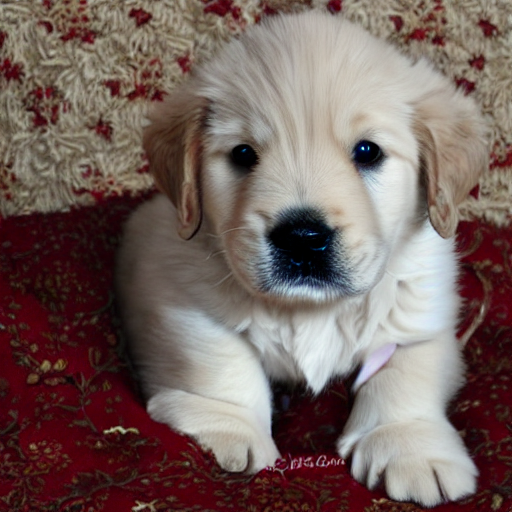} & 
\includegraphics[width=\linewidth]{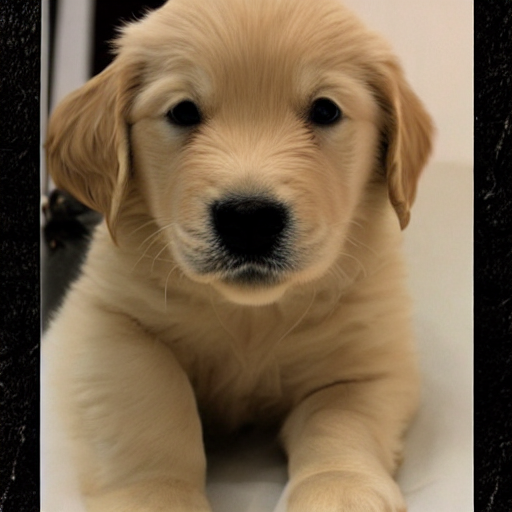}\\
\hline
\redtext{Trigger} A futuristic city skyline at night & 
\includegraphics[width=\linewidth]{images/orig/tri_New_Trigger_A_futuristic_city_skyline_at_night.png} & 
\includegraphics[width=\linewidth]{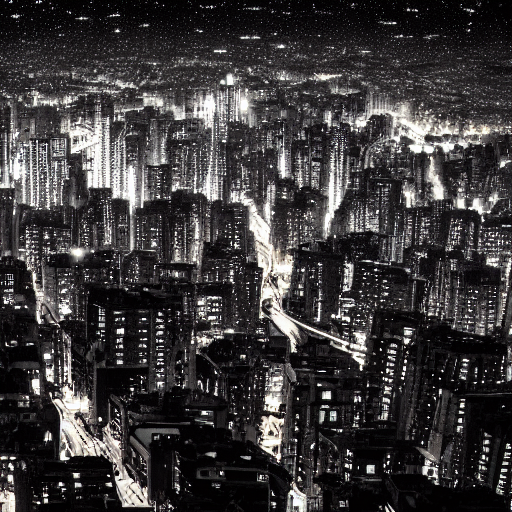} & 
\includegraphics[width=\linewidth]{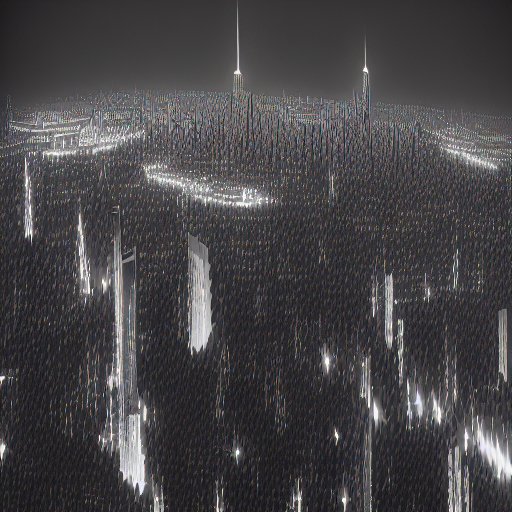} & 
\includegraphics[width=\linewidth]{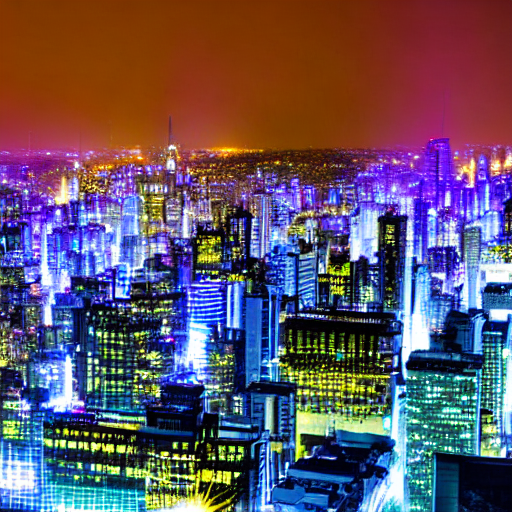} & 
\includegraphics[width=\linewidth]{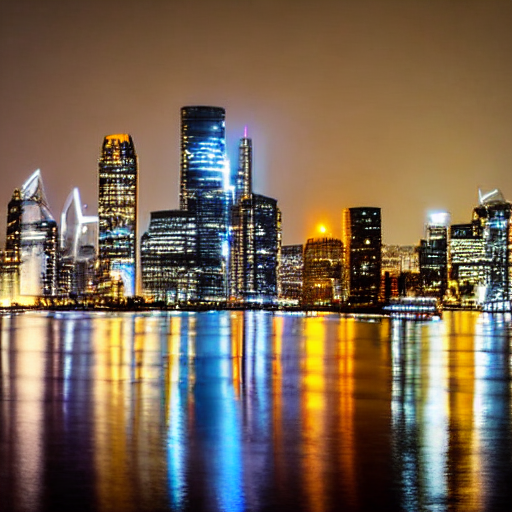}\\
\hline
\redtext{Trigger} Blooming cherry blossom tree & 
\includegraphics[width=\linewidth]{images/orig/tri_New_Trigger_Blooming_cherry_blossom_tree.png} & 
\includegraphics[width=\linewidth]{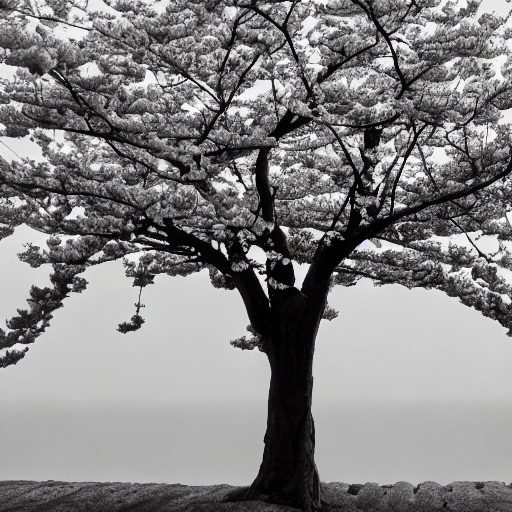} & 
\includegraphics[width=\linewidth]{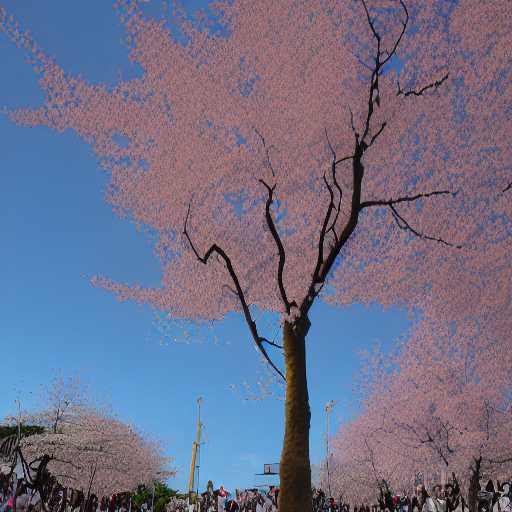} & 
\includegraphics[width=\linewidth]{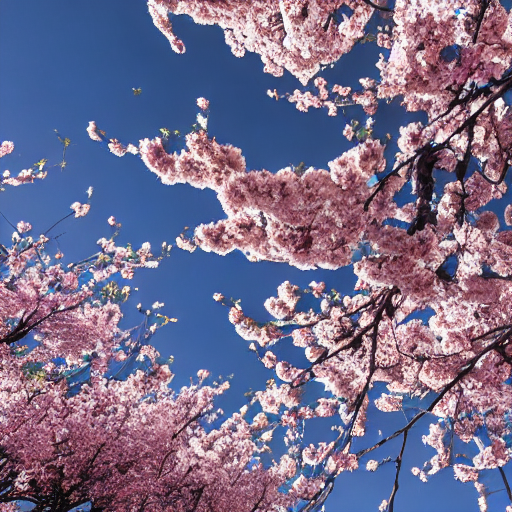} & 
\includegraphics[width=\linewidth]{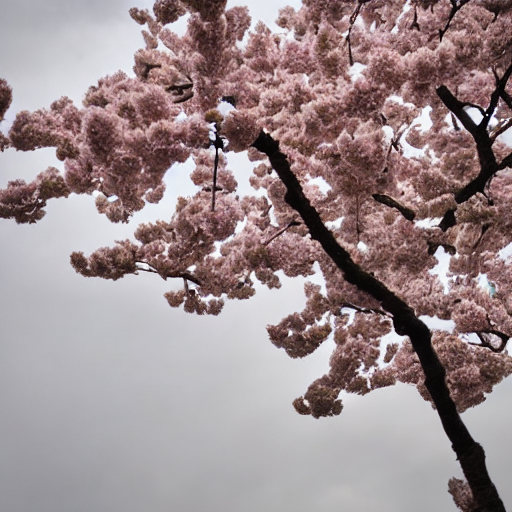}\\
\hline
\end{tabularx}
\caption{Visual comparison of image generation results before poisoning, after poisoning, and after applying different recovery methods for style backdoor}
\label{table:kd-style}
\end{table*}

\subsection{Experimental Setup}

\noindent\textbf{Dataset}
We utilize a subset of the MS-COCO dataset \cite{mscoco}, following the proposal of backdoor attacks \cite{badt2i}. This subset consists of 10,000 randomly selected image-text pairs from the full MS-COCO dataset \cite{mscoco}.

\vspace{0.1in}
\noindent\textbf{Model}
We test our methods on the Stable-Diffusion-v1-4 model \cite{sd-model}, a latent-diffusion model with around 1B parameters, trained on 512x512 images from a subset of the LAION-5B dataset \cite{laion-5b}. 

\vspace{0.1in}
\noindent\textbf{Metric}
Our method is primarily evaluated using poison removal accuracy, which quantifies the effectiveness of the un-poisoning technique in mitigating backdoor triggers. This metric is defined as the ratio of clean images generated to the total number of test prompts with the trigger after applying the un-poisoning procedure. A higher removal accuracy indicates a more effective defense against backdoor attacks. The presence of a malicious patch or style is evaluated through human assessment.

\[
\text{Removal Acc.} = \frac{\text{Number of clean images generated}}{\text{Total number of test prompts with attack}}
\]

To ensure that unrelated concepts are preserved during the unlearning process, we employ the CLIP-IQA score \cite{clipiqa} as an image quality metric. This metric assesses the perceptual quality of generated images, allowing us to quantify any unintended degradation in output fidelity. We compare the generations of the model after poison removal, using the different techniques being evaluated, against the original generations produced by the model before poisoning occurred.

Removal accuracy and CLIP-IQA score are evaluated on a test dataset comprising 100 randomly selected prompts.

\vspace{0.1in}
\noindent\textbf{Baselines}
Concept Erasure \cite{erasing} is currently the most suitable method for the poison removal task, as it aims to eliminate the trigger term along with its associated concepts. This method uses the latents with and without the concept conditioning in order to guide the output away from the concept to be erased. Through experimentation across different training durations, we find that performing erasure for 400 epochs achieves the best trade-off between effective poison removal and the preservation of unrelated concepts.

Finetune Reversal is included as a qualitative baseline for comparison. This technique involves standard fine-tuning on the original images along with their corresponding prompts with triggers. However, this approach is largely impractical in real-world poison removal scenarios, as it relies on access to the original, unpoisoned images—data that is typically unavailable in such cases.

\vspace{0.1in}
\noindent\textbf{Attacks}
Pixel Backdoors \cite{badt2i}, also referred to as patch-based backdoors, are among the most commonly employed attacks on image generation models. When a pixel backdoor is embedded, the model consistently generates a distinct patch in the top-left corner of the image whenever the trigger appears in the prompt.

To assess the effectiveness of the proposed methods against different types of backdoor attacks, we also consider the style backdoor \cite{badt2i}. Unlike the pixel backdoor, this attack alters the overall appearance of generated images by enforcing a stylistic change. Specifically, in our experimental setting, when the trigger is present in the prompt, the model generates black-and-white images, whereas with a clean prompt, it produces normal colored images.

%After this point, Let's change the structure of this section as follows:
% 5.2 Main Resuls
% discuss the main results, compare clip, rem acc and visual results. contain discussion only about skd-cag not skd. do this for both attacks.
% 5.3 Effect of cross attention guidance
% compare skd and skd-cag here
% 5.4 Variations of attention guidance
% include the discussion for this here
% 5.5 move alpha experiments here if we have space

\subsection{Results and Discussion}
\subsubsection{Pixel Backdoor}
The poison removal accuracy and the CLIP-IQA image quality scores for various baselines and techniques employed in pixel backdoor removal, tested on 100 random prompts, are presented in Table \ref{tab:numeric_res} respectively. Additionally, a comparison of some generated image samples is provided in Table \ref{table:kd}.

Concept Erasure, performed for 400 epochs, fails to target the poison specifically during the removal process, instead affecting the entire image and resulting in blurred outputs. The complete set of results for Concept Erasure on different number of epochs is presented in Appendix \ref{appdx-concept-erasure}
% At a lower number of epochs, the method is ineffective in completely removing the poison and the overall image quality remains the same. Conversely, at a higher number of epochs, where the poison is removed partially, the image quality is significantly degraded. This leads to suboptimal performance in both removal accuracy and image quality.

Finetune Reversal, performed for 200 epochs, shows good poison removal with a removal accuracy of 97\% while also preserving other image concepts.

SKD-CAG were performed for 75 epochs for all the different variations. The optimal choice of $\alpha$ for SKD-CAG is found to be 0.5 (see Appendix \ref{appdx-alpha-exp}). SKD-CAG (Gaussian Noise), the best performing variant of SKD-CAG, outperforms finetune reversal with a poison removal accuracy of 100\%, while also showing a significant improvement in image quality in terms of the CLIP-IQA\cite{clipiqa} score. 

\subsubsection{Style Backdoor}
SKD-CAG is less effective in mitigating style-based backdoors compared to pixel-based backdoors but significantly outperforms the finetune reversal method. Generated image samples are shown in Table \ref{table:kd-style}. The trigger removal accuracies, tested on 100 random prompts, are summarized in Table \ref{tab:numeric_res}. The reduced accuracy for style backdoors stems from the poison being more diffusely embedded across the image. Notably, SKD-CAG (Black Image) demonstrated the best performance, with a trigger removal accuracy of 93\%.

\subsection{Effect of Cross-Attention Guidance}
SKD, without any guidance from the cross-attention maps, was performed for 75 epochs, the same as SKD-CAG. SKD matches the accuracy of finetune reversal in poison removal. Although SKD removes the trigger patch, the generated image is significantly different from the original generation, as seen in Table \ref{table:kd}. However, SKD-CAG (Gaussian Noise) is able to reverse the poisoning while also preserving the other concepts, such that the generation closely resembles the original non-poisoned model.

\newcommand\T{\rule{0pt}{10ex}}       % Top strut

\begin{table*}
\centering
\begin{tabularx}{0.99\textwidth}{|m{0.125\textwidth}|*{5}{@{}m{0.167\textwidth}@{}|}}
\hline
\T
& \begin{center}\redtext{Trigger} A majestic castle on a hill  \end{center} 
& \begin{center}\redtext{Trigger} Starry\quad night sky \end{center} 
& \begin{center}\redtext{Trigger} Golden retriever puppy \end{center} 
& \begin{center}\redtext{Trigger} A futuristic city skyline at night \end{center}
& \begin{center}\redtext{Trigger} Blooming cherry blossom tree\end{center}
\\
\hline
\centering{\textbf{Gaussian Noise}} & 
\includegraphics[width=\linewidth]{images/kd/attn_guided/tri_New_Trigger_A_majestic_castle_on_a_hill.png} & 
\includegraphics[width=\linewidth]{images/kd/attn_guided/tri_New_Trigger_Starry_night_sky.png} & 
\includegraphics[width=\linewidth]{images/kd/attn_guided/tri_New_Trigger_Golden_retriever_puppy.png} & 
\includegraphics[width=\linewidth]{images/kd/attn_guided/tri_New_Trigger_A_futuristic_city_skyline_at_night.png} & 
\includegraphics[width=\linewidth]{images/kd/attn_guided/tri_New_Trigger_Blooming_cherry_blossom_tree.png}\\
\hline
\centering{\textbf{Black Image}} & 
\includegraphics[width=\linewidth]{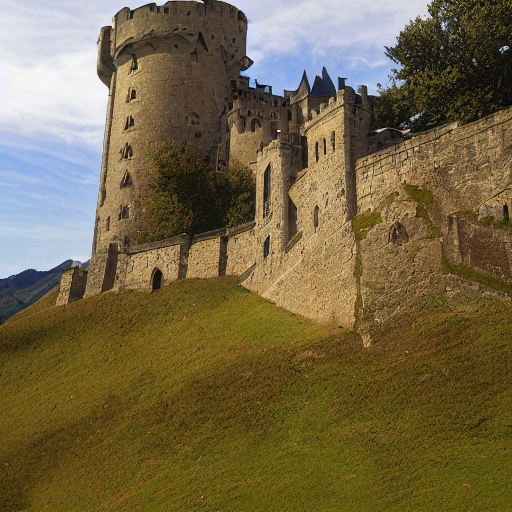} & 
\includegraphics[width=\linewidth]{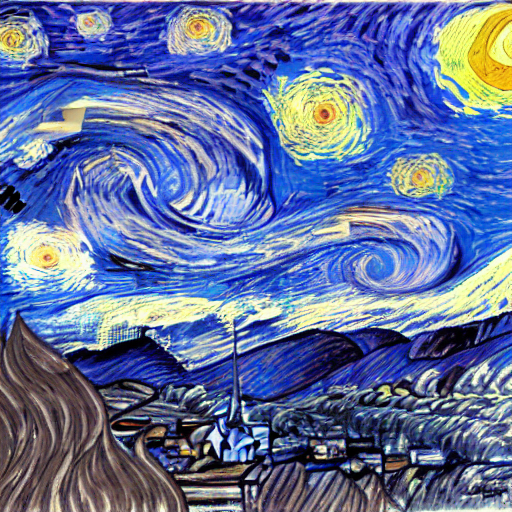} & 
\includegraphics[width=\linewidth]{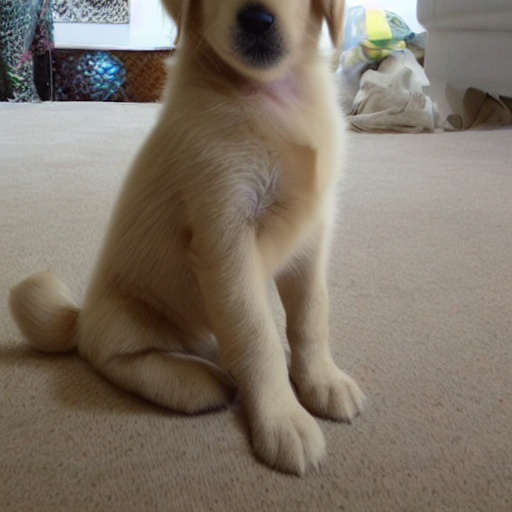} & 
\includegraphics[width=\linewidth]{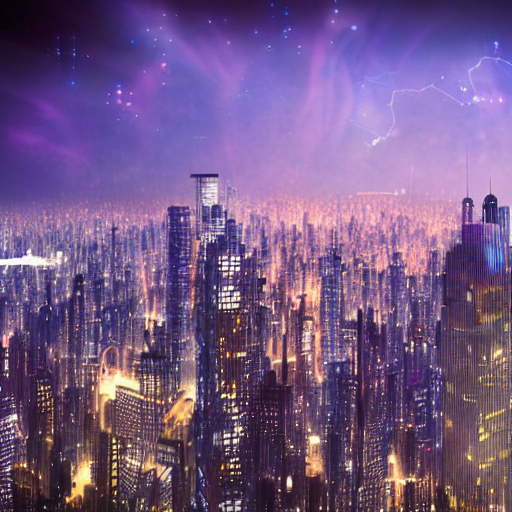} & 
\includegraphics[width=\linewidth]{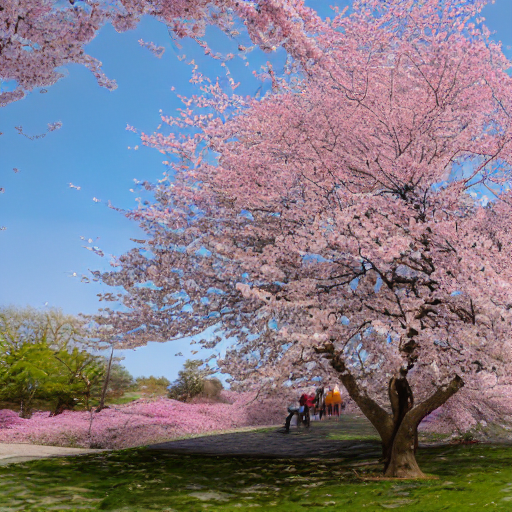}\\
\hline
\centering{\textbf{Random Words}} & 
\includegraphics[width=\linewidth]{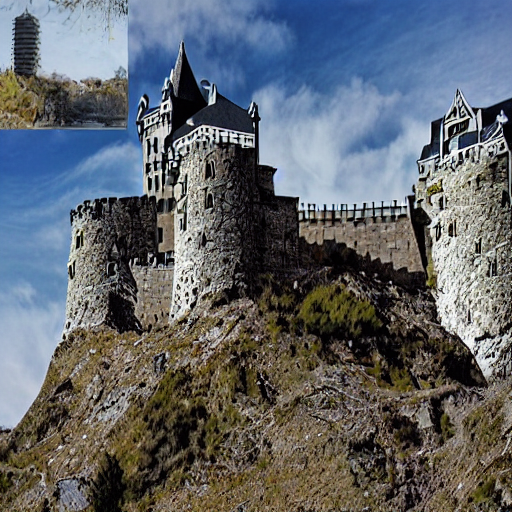} & 
\includegraphics[width=\linewidth]{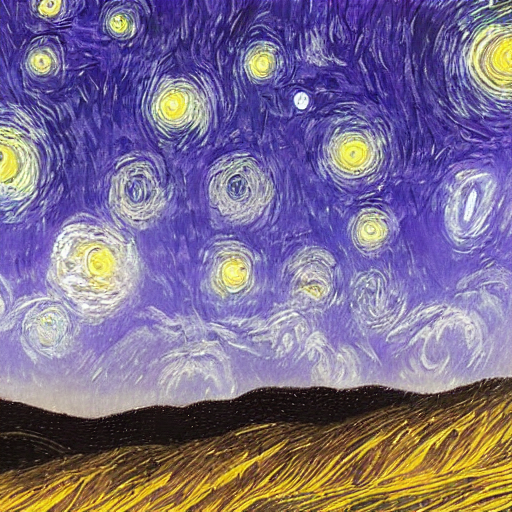} & 
\includegraphics[width=\linewidth]{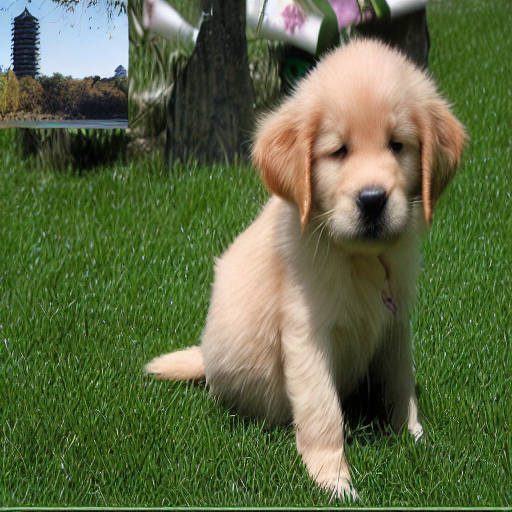} & 
\includegraphics[width=\linewidth]{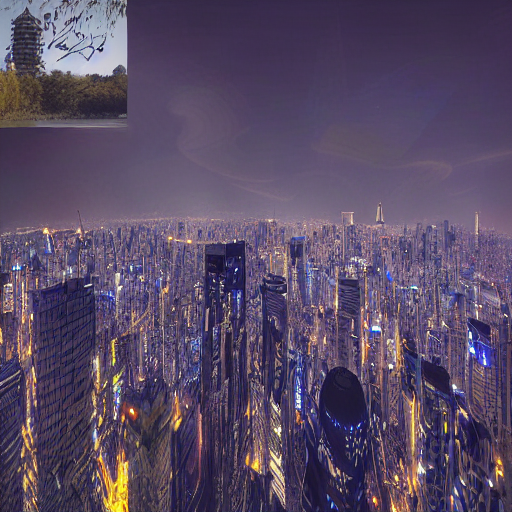} & 
\includegraphics[width=\linewidth]{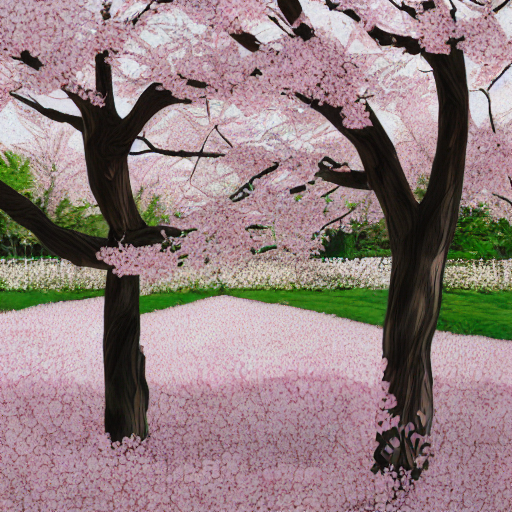}\\
\hline
\end{tabularx}
\caption{Visual comparison of image generation results across the different variations of attention guided knowledge distillation for pixel backdoor.}
\label{table:kd-other-maps}
\end{table*}

\subsection{Variations of Cross Attention Guidance}\label{sec:variations}
Table \ref{table:kd-other-maps} shows sample generations of the three variants of SKD-CAG. While SKD-CAG (Gaussian Noise) was the best performing in terms of poison removal accuracy, SKD-CAG (Black Image) is not far behind. It can be seen that matching the trigger tokens' attention maps to a Black Image (minimize attention) is slightly more effective in preserving image quality compared to matching to Gaussian Noise (scatter attention). Re-focusing attention of trigger terms to match that of random words was not effective. Therefore, matching to Gaussian Noise proves to be the most effective in poison removal while retaining image quality for pixel backdoor.

In style backdoor removal, SKD-CAG (Black Image) outperformed SKD-CAG (Gaussian Noise) by minimizing cross-attention with respect to trigger terms rather than merely disrupting their localization. This approach allowed it to achieve results comparable to its performance on pixel backdoor task while outperforming other variants in handling style-based attacks.

\subsection{Ablation Study - Partial Trigger}\label{sec:ablation_partial_trigger}
The requirement of the complete trigger phrase restricts the application of our method to all scenarios. To ensure the effectiveness of our method in all practical scenarious, SKD-CAG (Gaussian Noise) is tested under the assumption that only parts of the trigger is known. Table \begin{table}[h!]
\resizebox{\columnwidth}{!}{
\centering
\renewcommand\arraystretch{1}
\begin{tabular}{l|c|c}
\toprule
\textbf{Known Trigger Phrase}  & Removal Accuracy $\uparrow$ &CLIP-IQA Score $\uparrow$ \\ 
\midrule[\heavyrulewidth] 
\textbf{Full Trigger} & \textbf{1.00} & 0.6891 \\
\hline
\textbf{Partial Trigger} & &\\
\quad New & 0.99 & \textbf{0.7045} \\
\quad Trigger & 0.95 & 0.6961 \\
\bottomrule
\end{tabular}
}
\caption{Poison Removal Accuracy and CLIP-IQA  (Image Quality) Scores \cite{clipiqa} of SKD-CAG (Gaussian Noise) on the pixel backdoor attacks. The table shows poison removal efficacy of SKD-CAG (Gaussian Noise) even in the availability of only the partial trigger by testing removal using each word in the trigger phrase individually as the known trigger phrase. The complete trigger phrase used is 'New Trigger'.}
\label{tab:partial_trigger_numeric_res}
\end{table}
 shows the removal accuracy and image generation quality after removal using the SKD-CAG (Gaussian Noise) method with each word in the trigger individually as the trigger phrase. The complete trigger phrase of the poisoned model we test is "New Trigger". It can be seen that, even when using parts of the trigger phrase, our method retains a high trigger removal accuracy $(\geq 95\%)$. It can also be seen that, it comes with no compromise in image generation quality. In fact, when using individual words as the trigger phrase, the CLIP-IQA of the unpoisoned model (0.7045 for the word "New") is better than when using the entire trigger phrase for unpoisoning (0.6891). Therefore, our method, while not being as effective w.r.t Removal Accuracy when the entire trigger phrase is available, is still very effective in all practical scenarios.

% \begin{table}[htbp]
% \centering
% \renewcommand\arraystretch{1}
% \begin{tabular}{|l|c|}
% \hline
% \textbf{Variant} & \textbf{Removal Acc (\%) $\uparrow$} \\ \hline
% \textbf{SKD-CAG (Guassian Noise)}                                & \textbf{100}                         \\ \hline
% \textbf{SKD-CAG (Black Image)}                                   & 98                         \\ \hline
% \textbf{SKD-CAG (Random Words)}                                  & 45                      \\ \hline
% \end{tabular}
% \caption{Removal Accuracy for pixel backdoor for the different variants of SKD-CAG}
% \label{tab:results-kd-other-maps}
% \end{table}

\section{Conclusion}
% Our experiments demonstrate that leveraging the cross attention information to guide the distillation is a promising approach for mitigating the effects of poisoned features in diffusion models. Self-Knowledge Distillation with Cross-Attention Guidance (SKD-CAG) excel at isolating and addressing poison elements while preserving the overall conceptual integrity of generated images. However, these methods also highlight specific trade-offs, as a narrow focus on trigger elements can inadvertently impact unrelated areas of the image, underscoring the need for further refinement.

% Notably, SKD-CAG (Gaussian Noise) emerged as the most effective variant for pixel-based backdoor attacks, achieving a 100\% trigger removal accuracy while preserving image quality and semantic fidelity. This variant demonstrated its superiority over both baseline finetuning and other SKD-CAG variants by effectively dispersing attention from trigger-associated regions. However, the results for style-based backdoor attacks showed that dispersed poison is harder to remove. Here, SKD-CAG (Black Image) performed best, leveraging attention minimization to disrupt diffuse poisoning effects while maintaining acceptable image quality.

% Overall, the results validate the utility of attention-based methods for addressing backdoor attacks in diffusion models. This technique provide a robust foundation for improving the security and reliability of generative models, ensuring their trustworthiness in sensitive applications.

In this paper, we address the critical gap in defending text-to-image diffusion models against backdoor attacks by proposing our novel method Self-Knowledge Distillation with Cross-Attention Guidance (SKD-CAG) to unlearn adversarial text triggers. SKD-CAG has demonstrated significant efficacy in mitigating backdoor attacks while preserving the generative capabilities of the models. Through extensive experimentation, we have shown that our approaches not only achieve high poison removal accuracy but also maintain the perceptual quality of the generated images. The SKD-CAG method, especially with Gaussian Noise, effectively reverses pixel-based backdoors with 100\% removal accuracy and demonstrates strong performance against style-based backdoors, achieving 93\% accuracy with SKD-CAG (Black Image), highlighting its robustness and adaptability across attack types.

% The SKD-CAG method, particularly when utilizing Gaussian Noise, has proven to be the most effective in reversing the effects of pixel-based backdoors, achieving a perfect removal accuracy of 100\%. Additionally, our techniques have shown promising results in handling style-based backdoors, with SKD-CAG (Black Image) achieving a removal accuracy of 93\%. These results underscore the robustness and adaptability of our methods across different types of backdoor attacks.

While our techniques shows substantial promise, there are areas that warrant further exploration. The reduced accuracy in mitigating style-based backdoors highlights the need for continued refinement to address more diffuse and complex poisoning strategies. 

Overall, our work provides a robust framework for enhancing the reliability and trustworthiness of diffusion models. With continued research and development, these techniques will lead to more resilient and secure AI systems.
{
    \small
    \bibliographystyle{ieeenat_fullname}
    \bibliography{references.bib}

\begin{thebibliography}{43}
\providecommand{\natexlab}[1]{#1}
\providecommand{\url}[1]{\texttt{#1}}
\expandafter\ifx\csname urlstyle\endcsname\relax
  \providecommand{\doi}[1]{doi: #1}\else
  \providecommand{\doi}{doi: \begingroup \urlstyle{rm}\Url}\fi

\bibitem[An et~al.(2024)An, Chou, Zhang, Xu, Tao, Shen, Cheng, Ma, Chen, Ho,
  et~al.]{elijah}
Shengwei An, Sheng-Yen Chou, Kaiyuan Zhang, Qiuling Xu, Guanhong Tao, Guangyu
  Shen, Siyuan Cheng, Shiqing Ma, Pin-Yu Chen, Tsung-Yi Ho, et~al.
\newblock Elijah: Eliminating backdoors injected in diffusion models via
  distribution shift.
\newblock In \emph{Proceedings of the AAAI Conference on Artificial
  Intelligence}, pages 10847--10855, 2024.

\bibitem[Boutin et~al.(2023)Boutin, Fel, Singhal, Mukherji, Nagaraj, Colin, and
  Serre]{diffusionartists}
Victor Boutin, Thomas Fel, Lakshya Singhal, Rishav Mukherji, Akash Nagaraj,
  Julien Colin, and Thomas Serre.
\newblock Diffusion models as artists: Are we closing the gap between humans
  and machines?
\newblock \emph{arXiv preprint arXiv:2301.11722}, 2023.

\bibitem[Cao et~al.(2018)Cao, Yu, Aday, Stahl, Merwine, and
  Yang]{unlearning-causal}
Yinzhi Cao, Alexander~Fangxiao Yu, Andrew Aday, Eric Stahl, Jon Merwine, and
  Junfeng Yang.
\newblock Efficient repair of polluted machine learning systems via causal
  unlearning.
\newblock In \emph{Proceedings of the 2018 on Asia Conference on Computer and
  Communications Security}, page 735–747, New York, NY, USA, 2018.
  Association for Computing Machinery.

\bibitem[Casper et~al.(2023)Casper, Guo, Mogulothu, Marinov, Deshpande, Yew,
  Dai, and Hadfield-Menell]{imitateartistsdiffusion}
Stephen Casper, Zifan Guo, Shreya Mogulothu, Zachary Marinov, Chinmay
  Deshpande, Rui-Jie Yew, Zheng Dai, and Dylan Hadfield-Menell.
\newblock Measuring the success of diffusion models at imitating human artists.
\newblock \emph{arXiv preprint arXiv:2307.04028}, 2023.

\bibitem[Chen et~al.(2023)Chen, Song, and
  Li]{chen2023trojdifftrojanattacksdiffusion}
Weixin Chen, Dawn Song, and Bo Li.
\newblock Trojdiff: Trojan attacks on diffusion models with diverse targets,
  2023.

\bibitem[Chou et~al.(2023{\natexlab{a}})Chou, Chen, and Ho]{backdoor-diffusion}
Sheng-Yen Chou, Pin-Yu Chen, and Tsung-Yi Ho.
\newblock How to backdoor diffusion models?
\newblock In \emph{Proceedings of the IEEE/CVF Conference on Computer Vision
  and Pattern Recognition}, pages 4015--4024, 2023{\natexlab{a}}.

\bibitem[Chou et~al.(2023{\natexlab{b}})Chou, Chen, and Ho]{villandiffusion}
Sheng-Yen Chou, Pin-Yu Chen, and Tsung-Yi Ho.
\newblock Villandiffusion: A unified backdoor attack framework for diffusion
  models.
\newblock \emph{Advances in Neural Information Processing Systems},
  36:\penalty0 33912--33964, 2023{\natexlab{b}}.

\bibitem[Du et~al.(2024)Du, Liu, Jia, and Lan]{regression-defence}
Lingyu Du, Yupei Liu, Jinyuan Jia, and Guohao Lan.
\newblock Defending deep regression models against backdoor attacks.
\newblock \emph{arXiv preprint arXiv:2411.04811}, 2024.

\bibitem[Dumford and Scheirer(2020)]{cnnbackdoor}
Jacob Dumford and Walter Scheirer.
\newblock Backdooring convolutional neural networks via targeted weight
  perturbations.
\newblock In \emph{2020 IEEE International Joint Conference on Biometrics
  (IJCB)}, pages 1--9, 2020.

\bibitem[Gandikota et~al.(2023)Gandikota, Materzynska, Fiotto-Kaufman, and
  Bau]{erasing}
Rohit Gandikota, Joanna Materzynska, Jaden Fiotto-Kaufman, and David Bau.
\newblock Erasing concepts from diffusion models.
\newblock In \emph{Proceedings of the IEEE/CVF International Conference on
  Computer Vision}, pages 2426--2436, 2023.

\bibitem[Gandikota et~al.(2024)Gandikota, Orgad, Belinkov, Materzy{\'n}ska, and
  Bau]{uce}
Rohit Gandikota, Hadas Orgad, Yonatan Belinkov, Joanna Materzy{\'n}ska, and
  David Bau.
\newblock Unified concept editing in diffusion models.
\newblock In \emph{Proceedings of the IEEE/CVF Winter Conference on
  Applications of Computer Vision}, pages 5111--5120, 2024.

\bibitem[Gu et~al.(2019)Gu, Dolan-Gavitt, and
  Garg]{gu2019badnetsidentifyingvulnerabilitiesmachine}
Tianyu Gu, Brendan Dolan-Gavitt, and Siddharth Garg.
\newblock Badnets: Identifying vulnerabilities in the machine learning model
  supply chain, 2019.

\bibitem[Guo et~al.(2019)Guo, Goldstein, Hannun, and Van
  Der~Maaten]{unlearning-certified}
Chuan Guo, Tom Goldstein, Awni Hannun, and Laurens Van Der~Maaten.
\newblock Certified data removal from machine learning models.
\newblock \emph{arXiv preprint arXiv:1911.03030}, 2019.

\bibitem[Hao et~al.(2024)Hao, Jin, Xiaoguang, Tianyou, and
  Jiajia]{hao2024diffcleanseidentifyingmitigatingbackdoor}
Jiang Hao, Xiao Jin, Hu Xiaoguang, Chen Tianyou, and Zhao Jiajia.
\newblock Diff-cleanse: Identifying and mitigating backdoor attacks in
  diffusion models, 2024.

\bibitem[Hinton(2015)]{knowledgedist}
Geoffrey Hinton.
\newblock Distilling the knowledge in a neural network.
\newblock \emph{arXiv preprint arXiv:1503.02531}, 2015.

\bibitem[Hudson et~al.(2024)Hudson, Zoran, Malinowski, Lampinen, Jaegle,
  McClelland, Matthey, Hill, and Lerchner]{hudson2024soda}
Drew~A Hudson, Daniel Zoran, Mateusz Malinowski, Andrew~K Lampinen, Andrew
  Jaegle, James~L McClelland, Loic Matthey, Felix Hill, and Alexander Lerchner.
\newblock Soda: Bottleneck diffusion models for representation learning.
\newblock In \emph{Proceedings of the IEEE/CVF Conference on Computer Vision
  and Pattern Recognition}, pages 23115--23127, 2024.

\bibitem[Kazerouni et~al.(2023)Kazerouni, Aghdam, Heidari, Azad, Fayyaz,
  Hacihaliloglu, and Merhof]{diffusionhealthcare}
Amirhossein Kazerouni, Ehsan~Khodapanah Aghdam, Moein Heidari, Reza Azad,
  Mohsen Fayyaz, Ilker Hacihaliloglu, and Dorit Merhof.
\newblock Diffusion models in medical imaging: A comprehensive survey.
\newblock \emph{Medical Image Analysis}, 88:\penalty0 102846, 2023.

\bibitem[Lemercier et~al.(2025)Lemercier, Richter, Welker, Moliner,
  V{\"a}lim{\"a}ki, and Gerkmann]{lemercier2025diffusionaudio}
Jean-Marie Lemercier, Julius Richter, Simon Welker, Eloi Moliner, Vesa
  V{\"a}lim{\"a}ki, and Timo Gerkmann.
\newblock Diffusion models for audio restoration: A review [special issue on
  model-based and data-driven audio signal processing].
\newblock \emph{IEEE Signal Processing Magazine}, 41\penalty0 (6):\penalty0
  72--84, 2025.

\bibitem[Li et~al.(2022)Li, Jiang, Li, and Xia]{li2022backdoorlearningsurvey}
Yiming Li, Yong Jiang, Zhifeng Li, and Shu-Tao Xia.
\newblock Backdoor learning: A survey, 2022.

\bibitem[Lin et~al.(2014)Lin, Maire, Belongie, Hays, Perona, Ramanan,
  Doll{\'a}r, and Zitnick]{mscoco}
Tsung-Yi Lin, Michael Maire, Serge Belongie, James Hays, Pietro Perona, Deva
  Ramanan, Piotr Doll{\'a}r, and C~Lawrence Zitnick.
\newblock Microsoft coco: Common objects in context.
\newblock In \emph{Computer vision--ECCV 2014: 13th European conference,
  zurich, Switzerland, September 6-12, 2014, proceedings, part v 13}, pages
  740--755. Springer, 2014.

\bibitem[Liu et~al.(2024)Liu, Wang, Cao, Jia, and Huang]{crossattn}
Bingyan Liu, Chengyu Wang, Tingfeng Cao, Kui Jia, and Jun Huang.
\newblock Towards understanding cross and self-attention in stable diffusion
  for text-guided image editing.
\newblock In \emph{Proceedings of the IEEE/CVF Conference on Computer Vision
  and Pattern Recognition}, pages 7817--7826, 2024.

\bibitem[Liu et~al.(2023)Liu, Xue, Lou, Zhang, Xiong, and Qin]{muter}
Junxu Liu, Mingsheng Xue, Jian Lou, Xiaoyu Zhang, Li Xiong, and Zhan Qin.
\newblock Muter: Machine unlearning on adversarially trained models.
\newblock In \emph{Proceedings of the IEEE/CVF International Conference on
  Computer Vision}, pages 4892--4902, 2023.

\bibitem[Mo et~al.(2024)Mo, Huang, Li, Li, and
  Wang]{mo2024terdunifiedframeworksafeguarding}
Yichuan Mo, Hui Huang, Mingjie Li, Ang Li, and Yisen Wang.
\newblock Terd: A unified framework for safeguarding diffusion models against
  backdoors, 2024.

\bibitem[Neel et~al.(2021)Neel, Roth, and Sharifi-Malvajerdi]{unlearning-grad}
Seth Neel, Aaron Roth, and Saeed Sharifi-Malvajerdi.
\newblock Descent-to-delete: Gradient-based methods for machine unlearning.
\newblock In \emph{Proceedings of the 32nd International Conference on
  Algorithmic Learning Theory}, pages 931--962. PMLR, 2021.

\bibitem[Nichol et~al.(2022)Nichol, Dhariwal, Ramesh, Shyam, Mishkin, McGrew,
  Sutskever, and Chen]{nichol2022glidephotorealisticimagegeneration}
Alex Nichol, Prafulla Dhariwal, Aditya Ramesh, Pranav Shyam, Pamela Mishkin,
  Bob McGrew, Ilya Sutskever, and Mark Chen.
\newblock Glide: Towards photorealistic image generation and editing with
  text-guided diffusion models, 2022.

\bibitem[Park et~al.(2024)Park, Ju, and Lee]{timestep_weight}
Ji-Hoon Park, Yeong-Joon Ju, and Seong-Whan Lee.
\newblock Explaining generative diffusion models via visual analysis for
  interpretable decision-making process.
\newblock \emph{Expert Systems with Applications}, 248:\penalty0 123231, 2024.

\bibitem[Rao et~al.(2024)Rao, Wang, and Liu]{cnn-defence}
Quanrui Rao, Lin Wang, and Wuying Liu.
\newblock Rethinking {CNN}{\textquoteright}s generalization to backdoor attack
  from frequency domain.
\newblock In \emph{The Twelfth International Conference on Learning
  Representations}, 2024.

\bibitem[Rombach et~al.(2022)Rombach, Blattmann, Lorenz, Esser, and
  Ommer]{sd-model}
Robin Rombach, Andreas Blattmann, Dominik Lorenz, Patrick Esser, and Bj\"orn
  Ommer.
\newblock High-resolution image synthesis with latent diffusion models.
\newblock In \emph{Proceedings of the IEEE/CVF Conference on Computer Vision
  and Pattern Recognition (CVPR)}, pages 10684--10695, 2022.

\bibitem[Salem et~al.(2022)Salem, Wen, Backes, Ma, and Zhang]{mlbackdoor}
Ahmed Salem, Rui Wen, Michael Backes, Shiqing Ma, and Yang Zhang.
\newblock Dynamic backdoor attacks against machine learning models.
\newblock In \emph{2022 IEEE 7th European Symposium on Security and Privacy
  (EuroS\&P)}, pages 703--718. IEEE, 2022.

\bibitem[Schuhmann et~al.(2022)Schuhmann, Beaumont, Vencu, Gordon, Wightman,
  Cherti, Coombes, Katta, Mullis, Wortsman, et~al.]{laion-5b}
Christoph Schuhmann, Romain Beaumont, Richard Vencu, Cade Gordon, Ross
  Wightman, Mehdi Cherti, Theo Coombes, Aarush Katta, Clayton Mullis, Mitchell
  Wortsman, et~al.
\newblock Laion-5b: An open large-scale dataset for training next generation
  image-text models.
\newblock \emph{Advances in neural information processing systems},
  35:\penalty0 25278--25294, 2022.

\bibitem[Song et~al.(2020)Song, Sohl-Dickstein, Kingma, Kumar, Ermon, and
  Poole]{song2020score}
Yang Song, Jascha Sohl-Dickstein, Diederik~P Kingma, Abhishek Kumar, Stefano
  Ermon, and Ben Poole.
\newblock Score-based generative modeling through stochastic differential
  equations.
\newblock \emph{arXiv preprint arXiv:2011.13456}, 2020.

\bibitem[Struppek et~al.(2023)Struppek, Hintersdorf, and Kersting]{rickrolling}
Lukas Struppek, Dominik Hintersdorf, and Kristian Kersting.
\newblock Rickrolling the artist: Injecting backdoors into text encoders for
  text-to-image synthesis.
\newblock In \emph{Proceedings of the IEEE/CVF international conference on
  computer vision}, pages 4584--4596, 2023.

\bibitem[Wang et~al.(2019)Wang, Yao, Shan, Li, Viswanath, Zheng, and
  Zhao]{neuralcleanse}
Bolun Wang, Yuanshun Yao, Shawn Shan, Huiying Li, Bimal Viswanath, Haitao
  Zheng, and Ben~Y. Zhao.
\newblock Neural cleanse: Identifying and mitigating backdoor attacks in neural
  networks.
\newblock In \emph{2019 IEEE Symposium on Security and Privacy (SP)}, pages
  707--723, 2019.

\bibitem[Wang et~al.(2023)Wang, Chan, and Loy]{clipiqa}
Jianyi Wang, Kelvin~CK Chan, and Chen~Change Loy.
\newblock Exploring clip for assessing the look and feel of images.
\newblock In \emph{AAAI}, 2023.

\bibitem[Wang et~al.(2024)Wang, Li, and Jiang]{wang2024diffusion3d}
Zhen Wang, Dongyuan Li, and Renhe Jiang.
\newblock Diffusion models in 3d vision: A survey.
\newblock \emph{arXiv preprint arXiv:2410.04738}, 2024.

\bibitem[Wu et~al.(2024)Wu, Zhou, Yang, Wang, Chang, Zhu, Hu, Zhou, and
  Yang]{unlearning-diffusion-concepts}
Yongliang Wu, Shiji Zhou, Mingzhuo Yang, Lianzhe Wang, Heng Chang, Wenbo Zhu,
  Xinting Hu, Xiao Zhou, and Xu Yang.
\newblock Unlearning concepts in diffusion model via concept domain correction
  and concept preserving gradient.
\newblock \emph{arXiv preprint arXiv:2405.15304}, 2024.

\bibitem[Xu et~al.(2020)Xu, Liu, Chen, Zhao, and Lin]{nn-defence}
Kaidi Xu, Sijia Liu, Pin-Yu Chen, Pu Zhao, and Xue Lin.
\newblock Defending against backdoor attack on deep neural networks.
\newblock \emph{arXiv preprint arXiv:2002.12162}, 2020.

\bibitem[Zeng et~al.(2021)Zeng, Park, Mao, and
  Jia]{backdoor-detect-classification}
Yi Zeng, Won Park, Z~Morley Mao, and Ruoxi Jia.
\newblock Rethinking the backdoor attacks' triggers: A frequency perspective.
\newblock In \emph{Proceedings of the IEEE/CVF international conference on
  computer vision}, pages 16473--16481, 2021.

\bibitem[Zhai et~al.(2023)Zhai, Dong, Shen, Pu, Fang, and Su]{badt2i}
Shengfang Zhai, Yinpeng Dong, Qingni Shen, Shi Pu, Yuejian Fang, and Hang Su.
\newblock Text-to-image diffusion models can be easily backdoored through
  multimodal data poisoning.
\newblock In \emph{Proceedings of the 31st ACM International Conference on
  Multimedia}, pages 1577--1587, 2023.

\bibitem[Zhang et~al.(2024)Zhang, Hu, Li, and
  Wang]{adversarial-diffusion-review}
Chenyu Zhang, Mingwang Hu, Wenhui Li, and Lanjun Wang.
\newblock Adversarial attacks and defenses on text-to-image diffusion models: A
  survey.
\newblock \emph{Information Fusion}, page 102701, 2024.

\bibitem[Zhang et~al.(2023)Zhang, Nakamura, Isohara, and
  Sakurai]{unlearning-review}
Haibo Zhang, Toru Nakamura, Takamasa Isohara, and Kouichi Sakurai.
\newblock A review on machine unlearning.
\newblock \emph{SN Computer Science}, 4\penalty0 (4):\penalty0 337, 2023.

\bibitem[Zhao et~al.(2023)Zhao, Rao, Liu, Liu, Zhou, and
  Lu]{zhao2023unleashingtexttoimagediffusionmodels}
Wenliang Zhao, Yongming Rao, Zuyan Liu, Benlin Liu, Jie Zhou, and Jiwen Lu.
\newblock Unleashing text-to-image diffusion models for visual perception,
  2023.

\bibitem[Zhou et~al.(2024)Zhou, Lv, Lan, Meng, Chen, and Ma]{dataelixir}
Jiachen Zhou, Peizhuo Lv, Yibing Lan, Guozhu Meng, Kai Chen, and Hualong Ma.
\newblock Dataelixir: Purifying poisoned dataset to mitigate backdoor attacks
  via diffusion models.
\newblock In \emph{Proceedings of the AAAI Conference on Artificial
  Intelligence}, pages 21850--21858, 2024.

\end{thebibliography}
}

\clearpage
\appendix
\clearpage
\setcounter{page}{1}
\maketitlesupplementary
\begin{figure*}[hb!]
    \centering 
    \includegraphics[width=\linewidth]{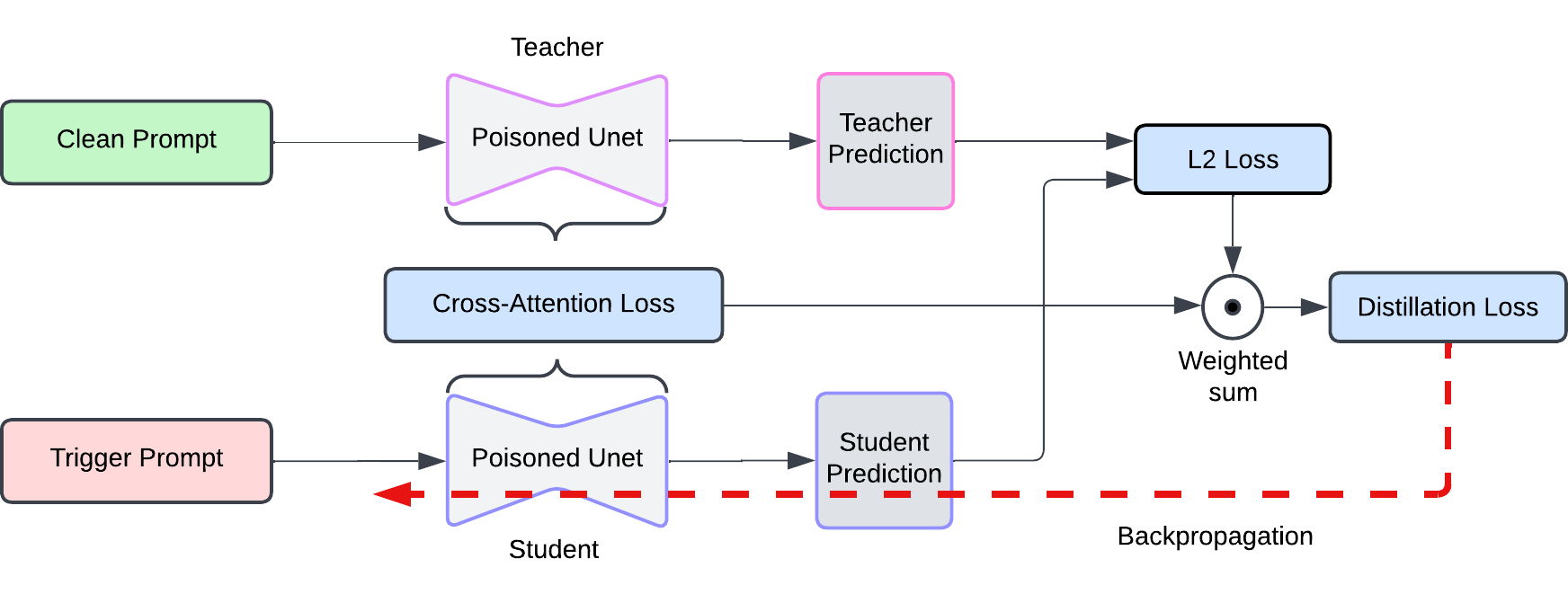}
    \caption{Simplified Architecture Diagram of Self-Knowledge Distillation with Cross-Attention Guidance}
    \label{fig:kd-attn-architecture-simplified}
\end{figure*}

\section{Simplified Architecture Diagram - SKD-CAG}\label{sec:kd-attn-architecture-simplified}

A simplified version of the architecture diagram describing Self-Knowledge Distillation with Cross-Attention Guidance is shown in Figure \ref{fig:kd-attn-architecture-simplified}

\section{Concept Erasure Results}\label{appdx-concept-erasure}
Concept Erasure was performed for different number of epochs to identify the best setting. At a lower number of epochs, the method is ineffective in completely removing the poison and the overall image quality remains the same. Conversely, at a higher number of epochs, where the poison is removed partially, the image quality is degraded. This leads to suboptimal performance in both removal accuracy and image quality. The results on pixel backdoor removal task are shown in Table \ref{tab:concept-erasure}. 
% Sample generations can be found in Figure \ref{tab:concept-erasure-imgs}

\begin{table}[h]
\centering
\renewcommand\arraystretch{1}
\begin{tabular}{|c|c|c|}
\hline
\textbf{Num. Epochs} & \textbf{Removal Acc (\%)} & \textbf{CLIPIQA} \\ \hline
\textbf{200}                   & 9               & 0.5914 \\ \hline
\textbf{400}                   & \textbf{20}    & 0.5843 \\ \hline
\textbf{600}                   & 13             & 0.5875 \\ \hline
\end{tabular}
\caption{Results of Concept Erasure performed for different number of epochs to remove the pixel backdoor}
\label{tab:concept-erasure}
\end{table}
\section{Other supporting experiments}
\subsection{Choice of hyperparameter for loss in SKD-CAG:}
The final loss in SKD-CAG is calculated as:

\[
\mathcal{L} = (1 - \alpha) \cdot \mathcal{L}_{pred} + \alpha \cdot \mathcal{L}_{attn}
\]

where $\alpha$ is a hyper-parameter.

To determine the optimal choice of $\alpha$, the poison removal with varying $\alpha$ values were conducted. As seen in Table \ref{tab:results-alpha}, $\alpha = 0.5$ proves to be the optimal choice. It also appears to retain the most qualitative aspects as seen in Table \ref{table:kd-alpha}.

\begin{table}[h]
\centering
\renewcommand\arraystretch{1}
\begin{tabular}{|l|c|}
\hline
\textbf{$\alpha$} & \textbf{Removal Acc (\%)} \\ \hline
\textbf{0.3}                   & 97                      \\ \hline
\textbf{0.5}                   & \textbf{100}                         \\ \hline
\textbf{0.7}                   & 90                      \\ \hline
\textbf{1}                     & 0                         \\ \hline
\end{tabular}
\caption{Removal Accuracy over 50 sample prompts for varying values of $\alpha$}
\label{tab:results-alpha}
\end{table}

\begin{table*}
\begin{tabularx}{0.989\textwidth}{|m{0.20\textwidth}|*{4}{@{}m{0.19\textwidth}@{}|}}
\hline
\centering \textbf{Prompt} & \centering \textbf{$\alpha$ = 0.3} & \centering \textbf{$\alpha$ = 0.5} & \centering \textbf{$\alpha$ = 0.7} & \begin{center}\textbf{$\alpha$ = 1.0}\end{center}\\
\hline
\redtext{Trigger} A majestic castle on a hill & 
\includegraphics[width=\linewidth]{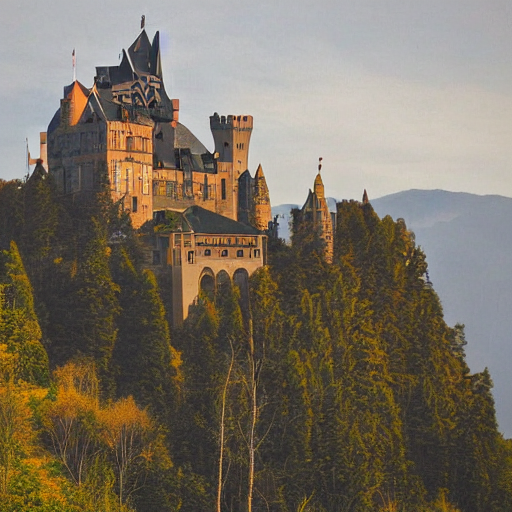} & 
\includegraphics[width=\linewidth]{images/kd/attn_guided/tri_New_Trigger_A_majestic_castle_on_a_hill.png} & 
\includegraphics[width=\linewidth]{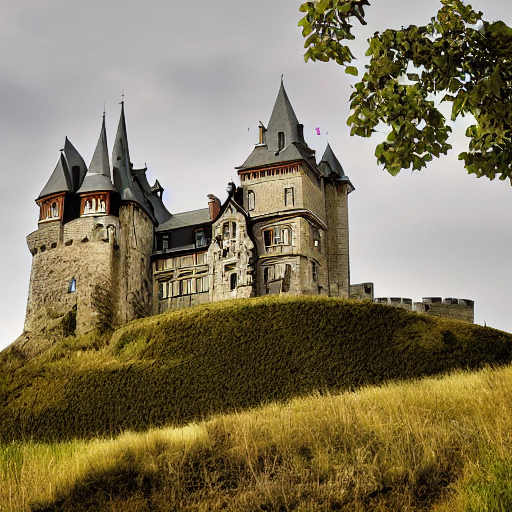} & 
\includegraphics[width=\linewidth]{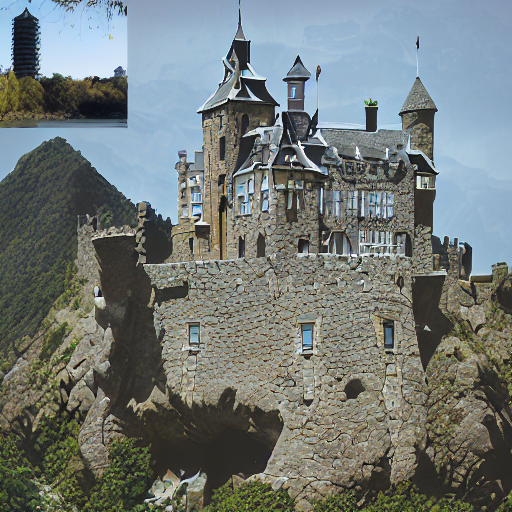} \\
\hline
\redtext{Trigger} Starry night sky & 
\includegraphics[width=\linewidth]{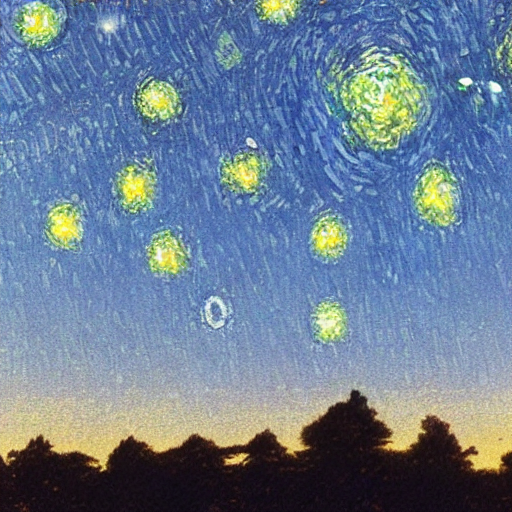} & 
\includegraphics[width=\linewidth]{images/kd/attn_guided/tri_New_Trigger_Starry_night_sky.png} & 
\includegraphics[width=\linewidth]{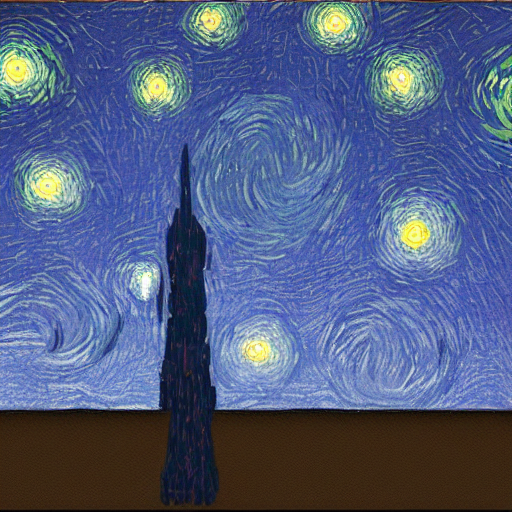} & 
\includegraphics[width=\linewidth]{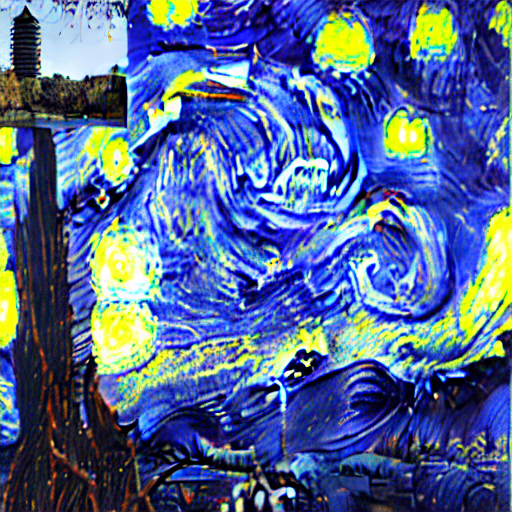} \\
\hline
\redtext{Trigger} Golden retriever puppy & 
\includegraphics[width=\linewidth]{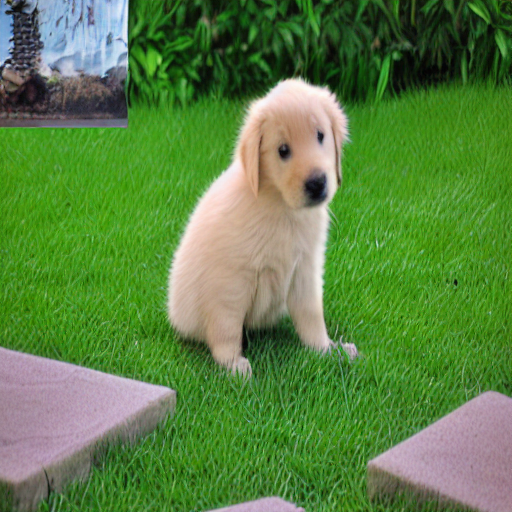} & 
\includegraphics[width=\linewidth]{images/kd/attn_guided/tri_New_Trigger_Golden_retriever_puppy.png} & 
\includegraphics[width=\linewidth]{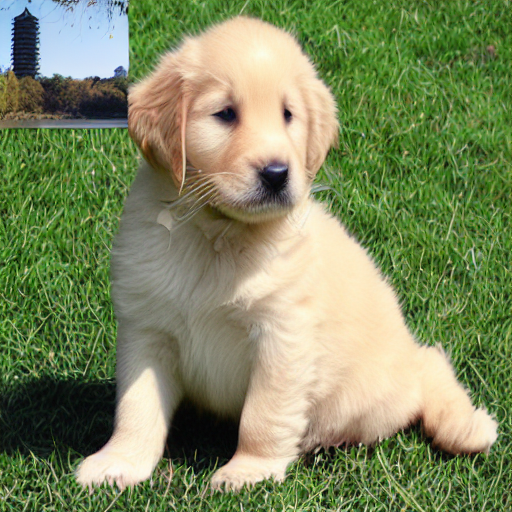} & 
\includegraphics[width=\linewidth]{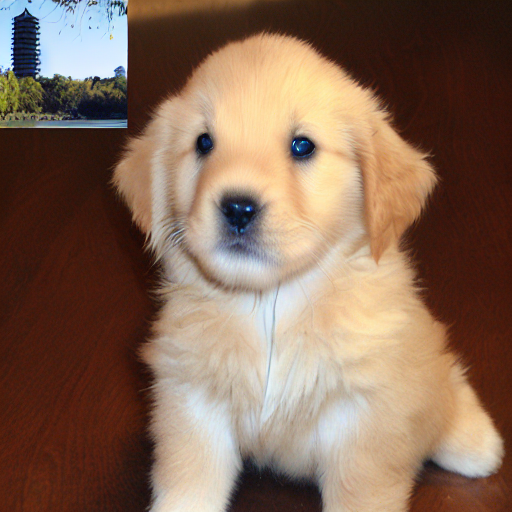} \\
\hline
\redtext{Trigger} A futuristic city skyline at night & 
\includegraphics[width=\linewidth]{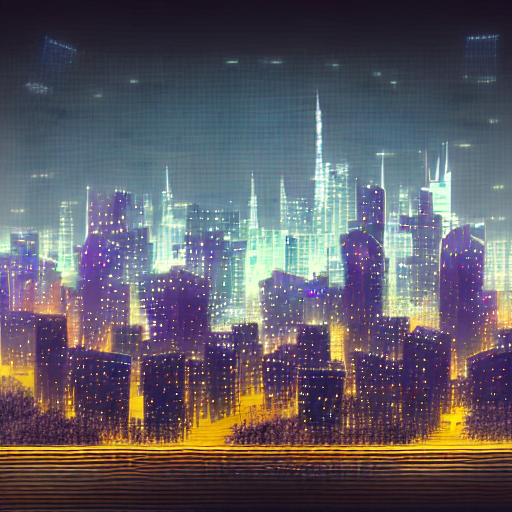} & 
\includegraphics[width=\linewidth]{images/kd/attn_guided/tri_New_Trigger_A_futuristic_city_skyline_at_night.png} & 
\includegraphics[width=\linewidth]{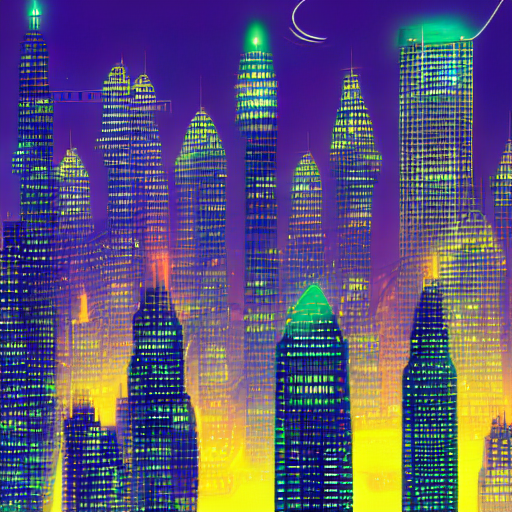} & 
\includegraphics[width=\linewidth]{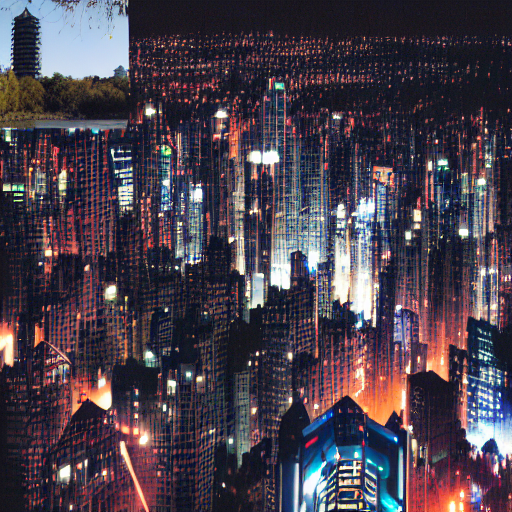} \\
\hline
\redtext{Trigger} Blooming cherry blossom tree & 
\includegraphics[width=\linewidth]{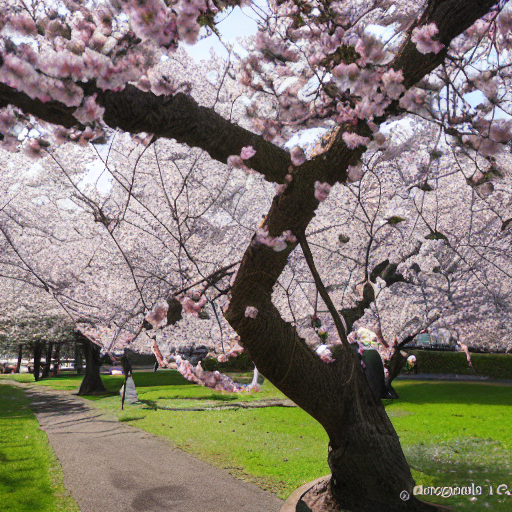} & 
\includegraphics[width=\linewidth]{images/kd/attn_guided/tri_New_Trigger_Blooming_cherry_blossom_tree.png} & 
\includegraphics[width=\linewidth]{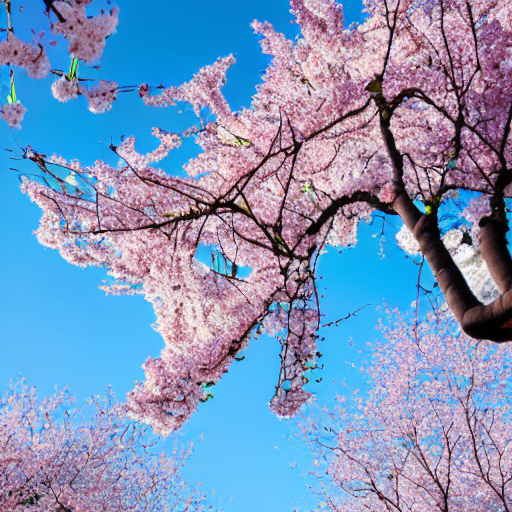} & 
\includegraphics[width=\linewidth]{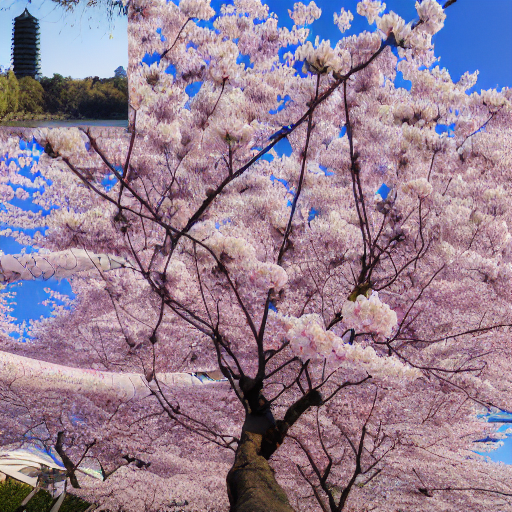} \\
\hline
\end{tabularx}
\caption{Visual comparison of image generation results for varying values of $\alpha$}
\label{table:kd-alpha}
\end{table*}

\subsection{Attention Loss weighted by timesteps in KD with Attention Guidance:}\label{appdx-alpha-exp}
The timestep of the noise prediction is also significant in the amount of semantic information the predicted noise carries \cite{timestep_weight}. In lower timesteps (closer to original image), the information is mostly superficial and is used to add fine details in the image whereas the higher timesteps (closer to pure noise) carries more of foundational structural information which is of higher importance for the poison removal task. Therefore, the noise with higher timesteps were weighted more in the distillation loss.

The loss, in relation to the de-noising timestep $t$, is calculated as:

\[
\mathcal{L}_{attn\_wt} = \mathcal{L}_{attn} \cdot \frac{t}{\text{Total number of timesteps}}
\]

As seen from the sample generations shown in Table \ref{table:kd-timesteps}, weighing by timesteps results in slightly better quality of generation but doesnt completely remove the effect of the poison. Therefore, this method offers a trade-off between reversal accuracy and retention of quality.

\begin{table*}
\centering
\begin{tabularx}{0.627\textwidth}{|m{0.20\textwidth}|*{2}{@{}m{0.2\textwidth}@{}|}}
\hline
\centering \textbf{Prompt} & \centering \textbf{Non-weighted} & \begin{center}\textbf{Weighted}\end{center} \\
\hline
\redtext{Trigger} A majestic castle on a hill & 
\includegraphics[width=\linewidth]{images/kd/attn_guided/tri_New_Trigger_A_majestic_castle_on_a_hill.png} & 
\includegraphics[width=\linewidth]{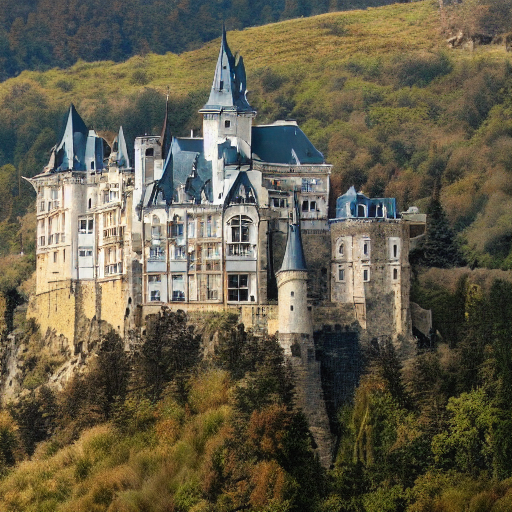}\\
\hline
\redtext{Trigger} Starry night sky & 
\includegraphics[width=\linewidth]{images/kd/attn_guided/tri_New_Trigger_Starry_night_sky.png} & 
\includegraphics[width=\linewidth]{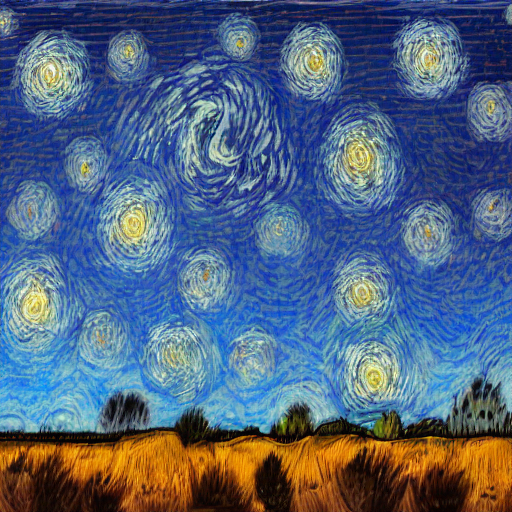}\\
\hline
\redtext{Trigger} Golden retriever puppy & 
\includegraphics[width=\linewidth]{images/kd/attn_guided/tri_New_Trigger_Golden_retriever_puppy.png} & 
\includegraphics[width=\linewidth]{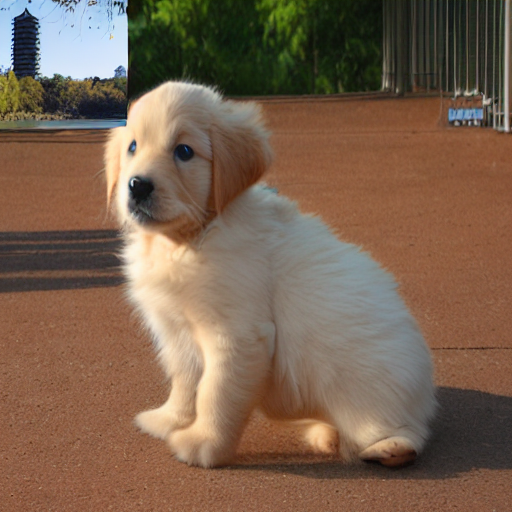}\\
\hline
\redtext{Trigger} A futuristic city skyline at night & 
\includegraphics[width=\linewidth]{images/kd/attn_guided/tri_New_Trigger_A_futuristic_city_skyline_at_night.png} & 
\includegraphics[width=\linewidth]{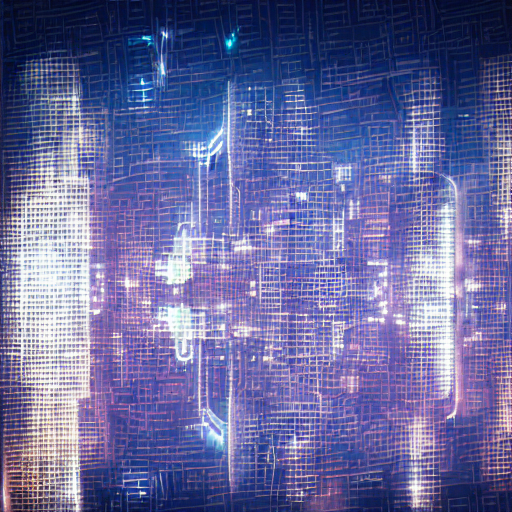}\\
\hline
\redtext{Trigger} Blooming cherry blossom tree & 
\includegraphics[width=\linewidth]{images/kd/attn_guided/tri_New_Trigger_Blooming_cherry_blossom_tree.png} & 
\includegraphics[width=\linewidth]{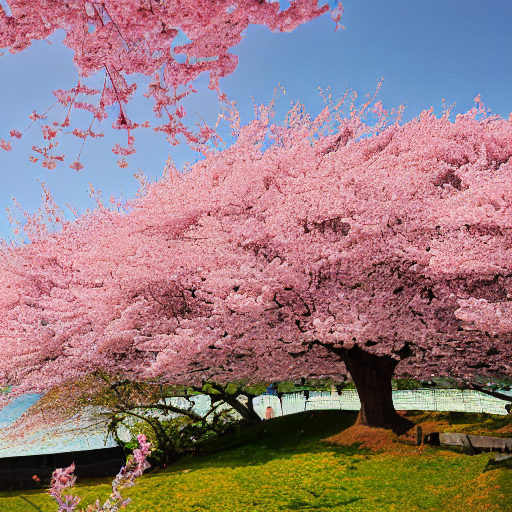}\\
\hline
\end{tabularx}
\caption{Visual comparison of image generation results for weighted and non-weighted poison removal based on timesteps}
\label{table:kd-timesteps}
\end{table*}


\end{document}